\documentclass[letterpaper, 10 pt, conference]{ieeeconf}  %
\IEEEoverridecommandlockouts                              
\usepackage{wrapfig}

\usepackage[colorlinks=true]{hyperref}
\hypersetup{colorlinks,breaklinks,
            urlcolor=[rgb]{1,0,0},
            citecolor=[rgb]{0,0,0.545}}

\usepackage{times}
\usepackage{xcolor}
\usepackage{comment}
\usepackage{cite}
\usepackage{soul}
\usepackage{graphicx}
\usepackage[linesnumbered,ruled,vlined]{algorithm2e}

\SetCommentSty{mycommfont}\usepackage[utf8]{inputenc}
\usepackage[small]{caption}
\usepackage{array}
\usepackage{amsmath}

\usepackage{multirow}

\title{\LARGE \bf Inferring 3D Shapes of Unknown Rigid Objects in Clutter through Inverse Physics Reasoning}

\author{
Changkyu Song and Abdeslam Boularias$^{1}$
\thanks{$^{1}$The authors are with the Department of Computer Science, Rutgers University, Piscataway, NJ, USA.
        {\tt\footnotesize \{cs1080, ab1544\}@cs.rutgers.edu}}%
}
\begin{document}

\markboth{IEEE Robotics and Automation Letters. Preprint Version. Accepted November, 2018}
{Song \MakeLowercase{\textit{et al.}}: Inferring 3D Shapes of Unknown Rigid Objects in Clutter through Inverse Physics Reasoning}

\maketitle

\begin{abstract}
We present a probabilistic approach for building, on the fly, $3$-D models of unknown objects while being manipulated by a robot. We specifically consider manipulation tasks in piles of clutter that contain previously unseen objects. 
Most manipulation algorithms for performing such tasks require known geometric models of the objects in order to grasp or rearrange them robustly. One of the  novel aspects of this work is the utilization of a physics engine for verifying hypothesized geometries in simulation. The evidence provided by physics simulations is used in a probabilistic framework that accounts for the fact that mechanical properties of the objects are uncertain. We present an efficient algorithm for inferring occluded parts of objects based on their observed motions and mutual interactions. Experiments using a robot show that this  approach is efficient for constructing physically realistic $3$-D models, which can be useful for manipulation planning. Experiments also show that the proposed approach significantly outperforms alternative approaches in terms of shape accuracy.
\end{abstract}
\section{Introduction}
\begin{wrapfigure}{tr}{0.25\textwidth}
\centering
\includegraphics[width=0.25\textwidth]{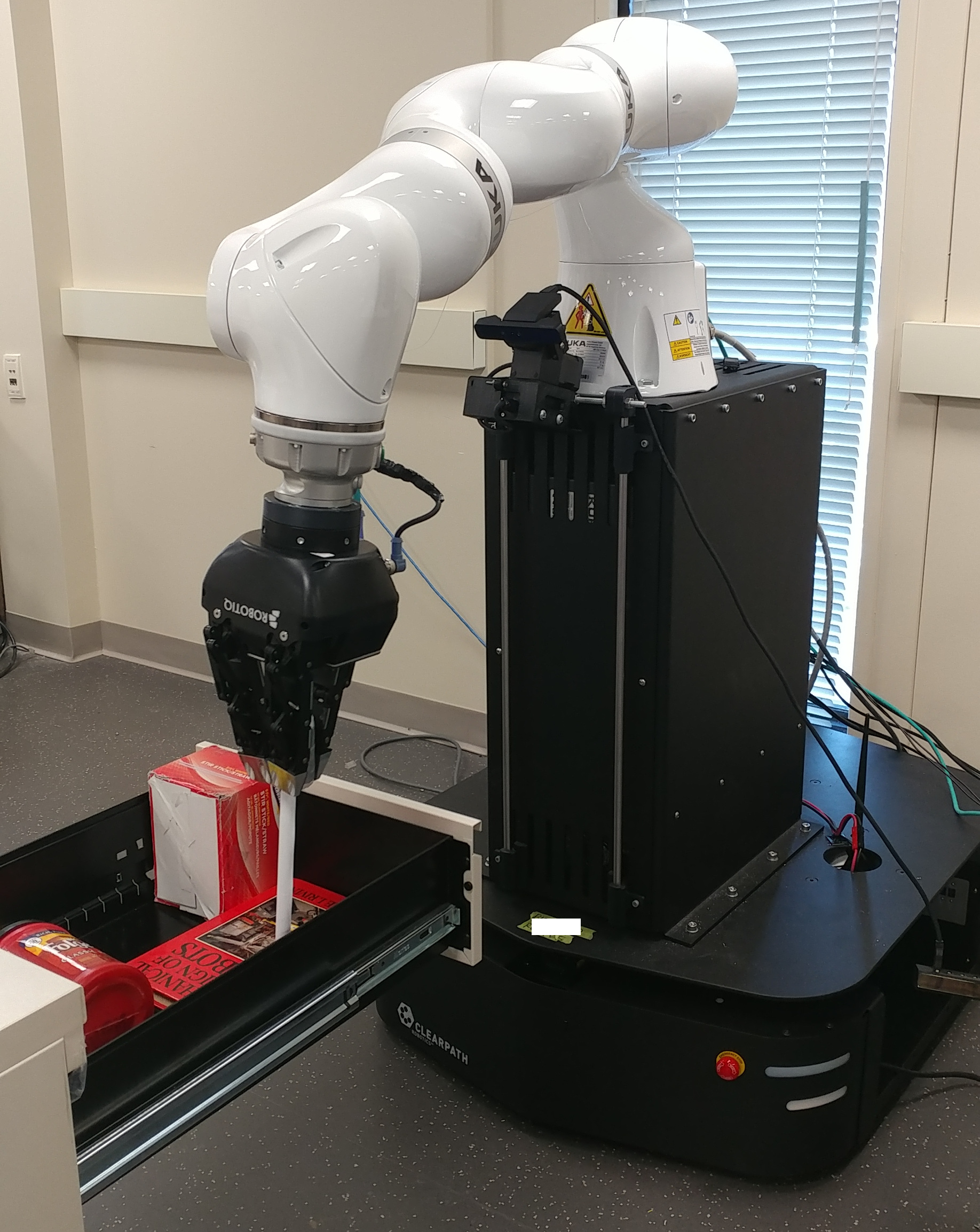}
\caption{\small {Experiments are performed using a {\it Kuka} arm mounted on a {\it Clearpath} mobile platform and equipped with a {\it Robotiq} hand and a depth-sensing camera {\it SR300}. }}
\vspace{-0.3cm}
\label{fig:kuka}
\end{wrapfigure}

Primates learn to manipulate all types of unknown objects from an early age. Yet, this seemingly trivial capability is still a major challenge when it comes to robots~\cite{specialissuemanipulation2013,Bohg}. Consider for instance the task of searching for an object inside a drawer, as illustrated in Figure~\ref{fig:kuka}.  To perform this task, the robot needs to detect the objects in the scene, and to plan grasping, pushing, and poking actions that would reveal the position of the searched object. 
The majority of motion planning algorithms, such as RRT and PRM~\cite{LaValle:2006:PA:1213331}, require geometric models of the objects involved in the task. The need for models has been put on display particularly during the {\it Amazon Picking Challenge}~\cite{7583659}, where robots were tasked with retrieving objects from narrow shelves, and collisions of the picked objects with other objects were a major source of failure, due to inaccurate estimates of the objects' poses.

In warehouses and factories, manipulated objects are typically known in advance, with their CAD models obtained from full 3D scans~\cite{RennieSBS16,Callietal_RAM2015,Janoch2011,furrer2017}.
Recent research efforts in grasping and manipulation are focused rather on tasks where object models are unavailable~\cite{6696928,SungLS17,PintoG16,KroemerJRAS_66360,detry2017c}. 
While most of these new methods ignore object modeling all together and focus on learning actions directly, other works have also explored automated modeling of unknown 3D objects~\cite{4543223}. A common approach consists in taking point clouds from multiple views and merging them using the popular Iterative Closest Point (ICP) technique~\cite{Bouaziz:2013,fusionpp2018}.
A large body of related works, known as {\it active vision}, is concerned with selecting the point of view of the camera to maximize
information gain with respect to the location of an object~\cite{KahnSPBRGA15,KraininCF11,LPKTLPICRA12}.
There is also a growing interest in robotics on {\it interactive perception}, wherein a manipulator intervenes on the scene by pushing certain objects so as to improve segmentation or object recognition~\cite{6631288,8007233,4543220,hoof2014probabilistic}. Our approach differs form these works in two aspects. First, our goal is to construct full CAD models that can be used by manipulation planning algorithms, and not to improve segmentation or object recognition. Second, we are concerned here only with predicting shapes of manipulated objects from RGB-D images, and not with optimizing the data collection process, which can be achieved by combining our approach with techniques for selecting camera views or poking/pushing actions. In this work, the camera is fixed and the objects pushed by the robot are chosen randomly.
\begin{figure*}[t]
    \centering 
\includegraphics[width=1.0\textwidth]{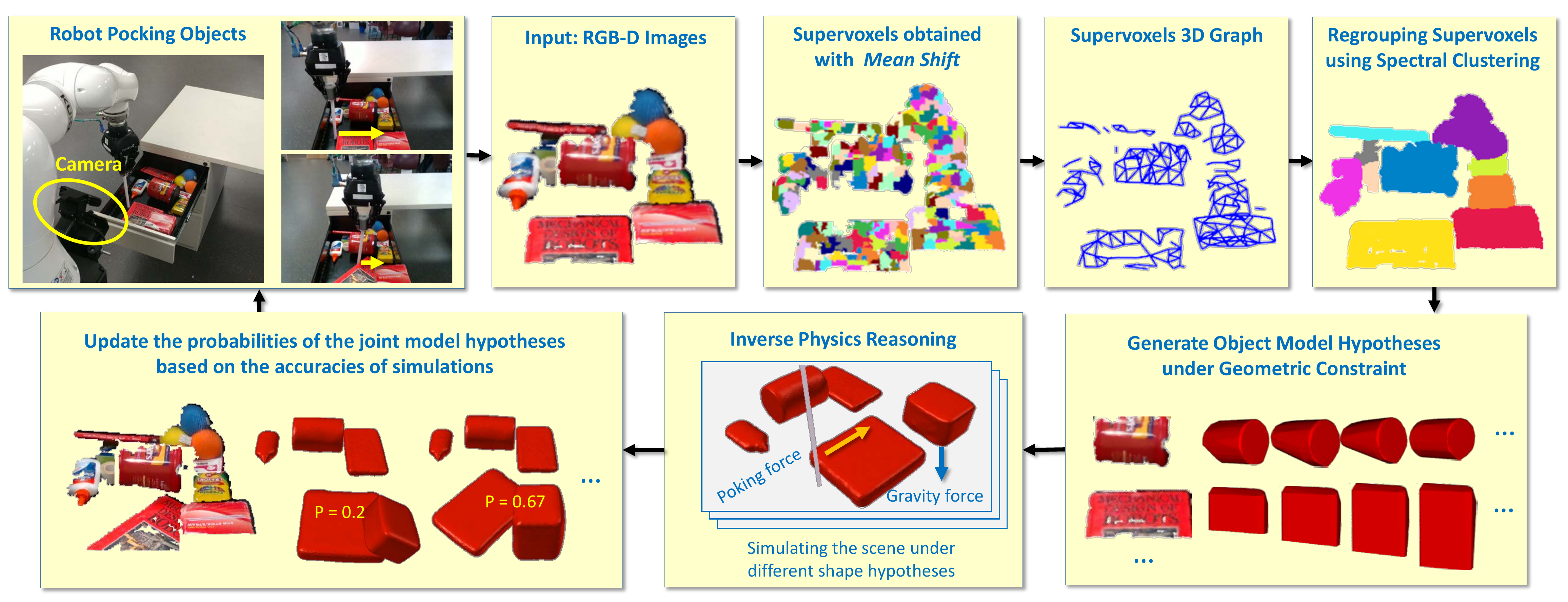}
\caption{Work-flow of the integrated system}
\vspace{-0.5cm}
\label{fig:pipeline}
\end{figure*}

{\it Volumetric shape completion} for partially occluded objects is an increasingly popular topic in computer vision~\cite{7298863,DaiQN17,VarleyDRRA17}. Learning-based approaches typically focus on known objects or specific categories, such as furniture~\cite{WuSKYZTX15,jrock-cvpr-2015,firman-cvpr-2016,dai2017complete}.
Approaches for unknown objects use energy minimizing solutions that penalize curvature variation~\cite{Kimia2003}, extract geometric primitives (planes or cylinders) from 3D meshes~\cite{Attene2006}, or exploit symmetry and {\it Manhattan properties}~\cite{gao2016exploiting}.
Some works have also considered {\it physical reasoning} for shape completion. For instance,\cite{Zheng_CVPR13, smzkxzm_imaginingTheUnseen_sigga14} presented an approach for scene understanding by reasoning about the physical stability of objects in a point cloud. Our method differs by its use of a physics engine to simulate both a robot's action and the gravitational and normal forces exerted upon a pile of objects, in addition to probabilistically reasoning about the unknown mechanical properties, and visually tracking the objects being pushed. This approach is inspired from previous works in cognitive science that have shown that knowledge of intuitive Newtonian principles and probabilistic representations are important for human-level complex scene understanding~\cite{DBLP:conf/cogsci/HamrickBT11,Battaglia18327}. Note also that there are works that use physical reasoning to predict the stability of a scene from an image~\cite{0003LF17}. We are interested in the inverse problem here, i.e predicting shapes of objects based on observed motions or stability of a scene. 

In this paper, we present an integrated system that combines: a robotic manipulator for pushing/poking objects in clutter, a segmentation and clustering module that detects objects from RGB-D images, and an inverse physical reasoning unit that infers missing parts of objects by replaying the robot's actions in simulation using multiple hypothesized shapes and assigning higher probabilities to hypotheses that better match the observed RGB-D images.
A video of the experiments along with a dataset containing annotated robotic actions and ground-truth 3D models and 6D poses of objects are  available at \url{https://goo.gl/1oYLB7}. 

\section{Overview of the Proposed Method}
A high-level overview of the proposed system is illustrated in Figure~\ref{fig:pipeline}. 
The system takes as inputs a sequence of RGB-D images of a clutter as well as recorded pushing or poking actions performed by a robot, and returns complete 3D models of the objects in the clutter. The system proceeds by first segmenting and clustering the given point clouds into objects. The parts of the objects that are hidden are hypothesized and sampled from a spectrum of possibilities. Each hypothesized object model is assigned a probability. The system then proceeds by replaying the robot's actions using various hypothesized object models, and comparing the movements of the objects in simulation to their observed real motions. The probabilities of the models that result in the most realistic simulations are systematically increased by using the reality gap as a likelihood function.

\section{Scene Segmentation}
\label{sec:seg_and_tracking}
\subsection{Segmentation}
RGB-D images of the clutter scene are obtained from a depth camera and is segmented as follows. We start by removing the known planes (tabletops and containers) using the RANSAC method. The robot's arm and hand are also removed from the point cloud using a known model of the robot and the corresponding forward kinematics. Each point cloud is segmented into a set of supervoxels by using the {\it mean shift} algorithm. A supervoxel is a small cluster of 3D points that share the same color. Then, a graph of supevoxels is created by connecting pairs of supevoxels that share a boundary in the corresponding point cloud. The edges connecting supervoxels are weighted according to the directions of their average surface normals, as proposed in~\cite{Stein2014}. A convexity prior is enforced here, by assigning smaller weights to edges that connect concave surfaces.
An edge $(i,j)$ is weighted with $w_{i,j} = \max\{v^t_i.(c_i-c_j) , v^t_j.(c_j-c_i) , 0 \}$, where $c_i$ and $c_j$ are the 3D centers of adjacent supervoxels $i$ and $j$ respectively, $v_i$ and $v_j$ are their respective surface normals. 
Using the {\it spectral clustering} technique~\cite{NIPS2001:2092}, the supervoxels are clustered into objects based on the weights of their connections. Namely, the normalized Laplacian $L_{sym}$ of the weighted adjacency matrix of the graph is computed, and the first $n$ eigenvectors of $L_{sym}$ are retained. $n$ is automatically determined by ranking the eigen values and cutting off at the first value that significantly differs from the others. Finally, the objects are obtained by clustering the supervoxels according to their coordinates in the retained eigenvectors, using the {\it k-means} algorithm. 
Thanks to this hierarchical approach, we reduced the running time of the spectral clustering layer by orders of magnitude. For example, segmenting the scenes shown in Figure~\ref{fig:pipeline} required about ten milliseconds on a single CPU. 

\subsection{Facet Decomposition}
\begin{figure}[h]
\centering
\includegraphics[width=0.18\textwidth,trim={0 0 0 0},clip]{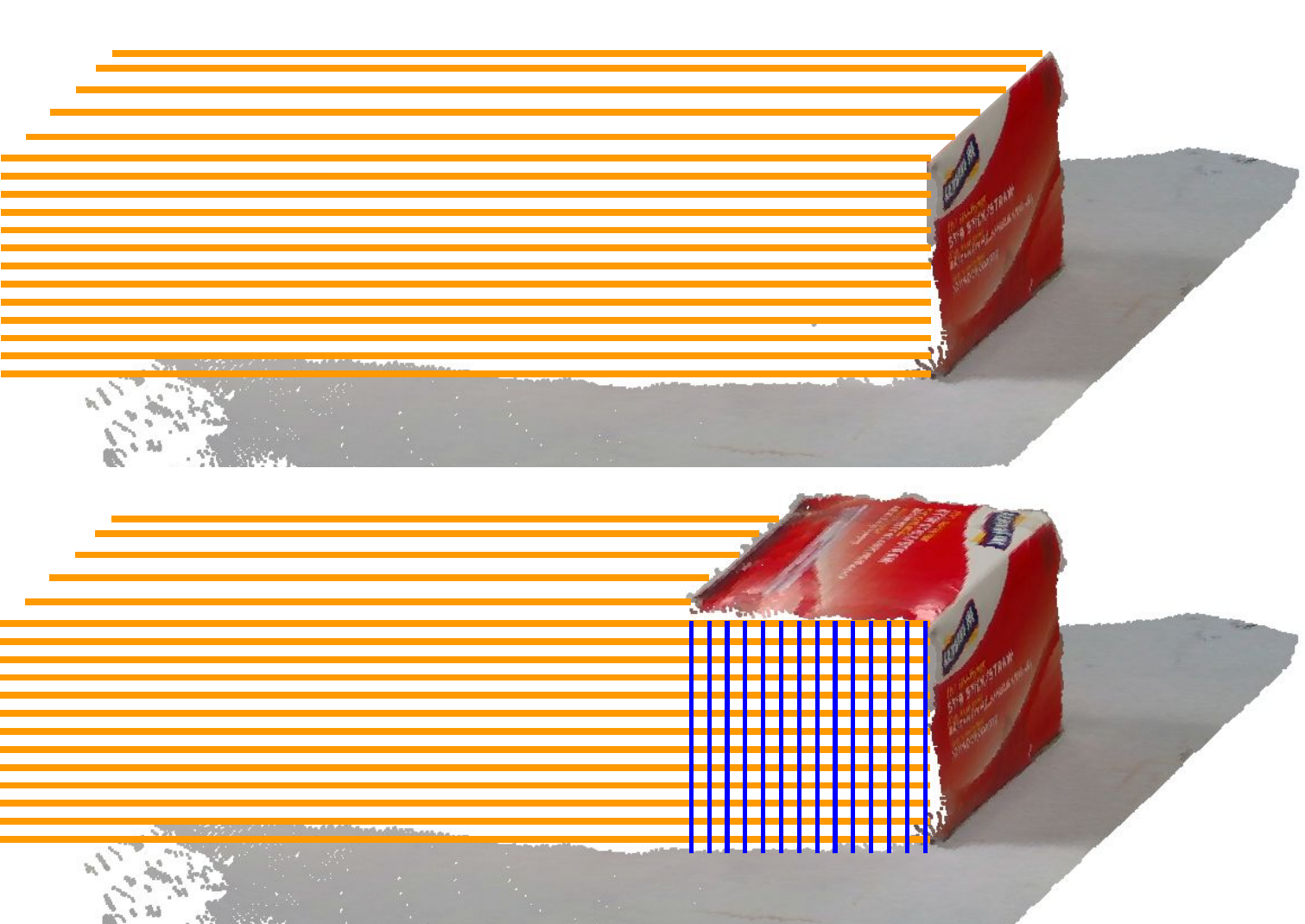} 
\includegraphics[width=0.18\textwidth,trim={0 0 0 6cm},clip]{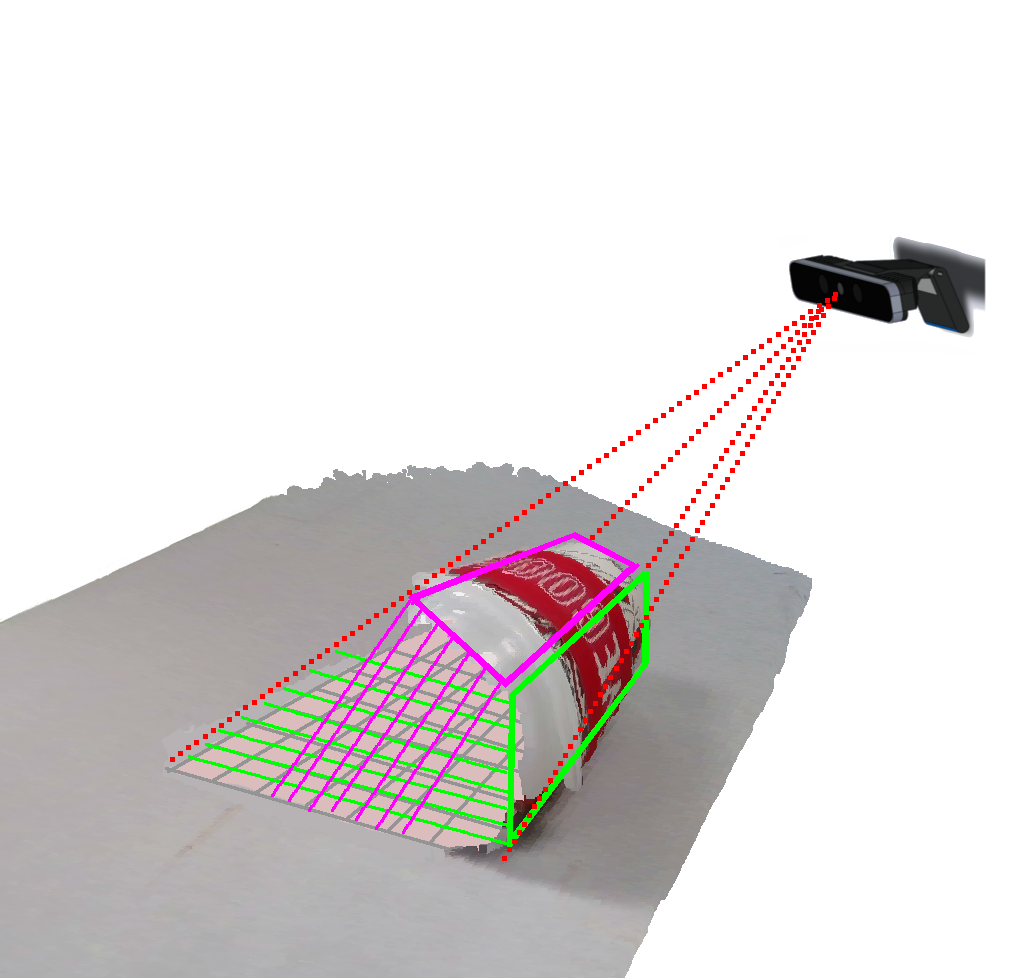}
\caption{Observed facets, and domains of potential hidden facets}
\label{fig:camera_ray}
\end{figure}

The result of segmentation and tracking process is a set of $n$ partial objects, $\{O_1, O_2, \dots, O_n\}$, wherein each partial object $O_i$ is a set of {\it facets}, i.e. $O_i = \{F^o_1, F^o_2, \dots F^o_k\}$. A facet is a  small homogeneous region that belongs to a side of an object. For instance, a cubic object is made of six facets, whereas a spherical object can be approximately modeled as a large set of small facets. The facets of an object are obtained by clustering its supervoxels into larger regions, using the curvature calculated from the normals as a distance in the mean shift algorithm. Figure~\ref{fig:camera_ray} shows simple examples of partial objects segmented into facets using this process.
\vspace{-0.3cm}

\section{Inverse Physics Reasoning}
The objective of the inverse physics reasoning is the inference of plausible full models that complete the observed partial models of objects $\{O_i\}_{i=1}^{n}$, by simulating the forces applied on the objects by the robot and environment and weighing the hypothesized models based on how accurately they predict the observations. We start by describing the range of shapes considered here, then we formulate the inference problem, and present our solution to the problem.

\subsection{Probabilistic Object Models}
We define an object model $X_i$ as a set of facets $\{F_1, F_2, \dots F_m\}$, wherein each facet is itself a set of 3D points in a common coordinate system. A partial object $O_i$ is a set of observed facets that belong to $X_i$, i.e. $O_i \subseteq X_i$. Therefore, an object model is the union of two sets of facets, observed ones and hypothesized unseen ones, i.e. $X_i = O_i \cup H_i$ where $H_i = \{F^h_j\}_{j=1}^{l}$ is the set of imagined hidden facets. We define $P(X_i)$ as the probability that the object with observed facets $O_i$ has exactly $l$ additional hidden facets given in $H_i = X_i - O_i$. Our goal is to estimate $P(X_i)$.

\subsection{Facet Hypotheses}
\begin{wrapfigure}{r}{0in}
\centering
\hspace{-0.2cm}
\includegraphics[width=0.25\textwidth]{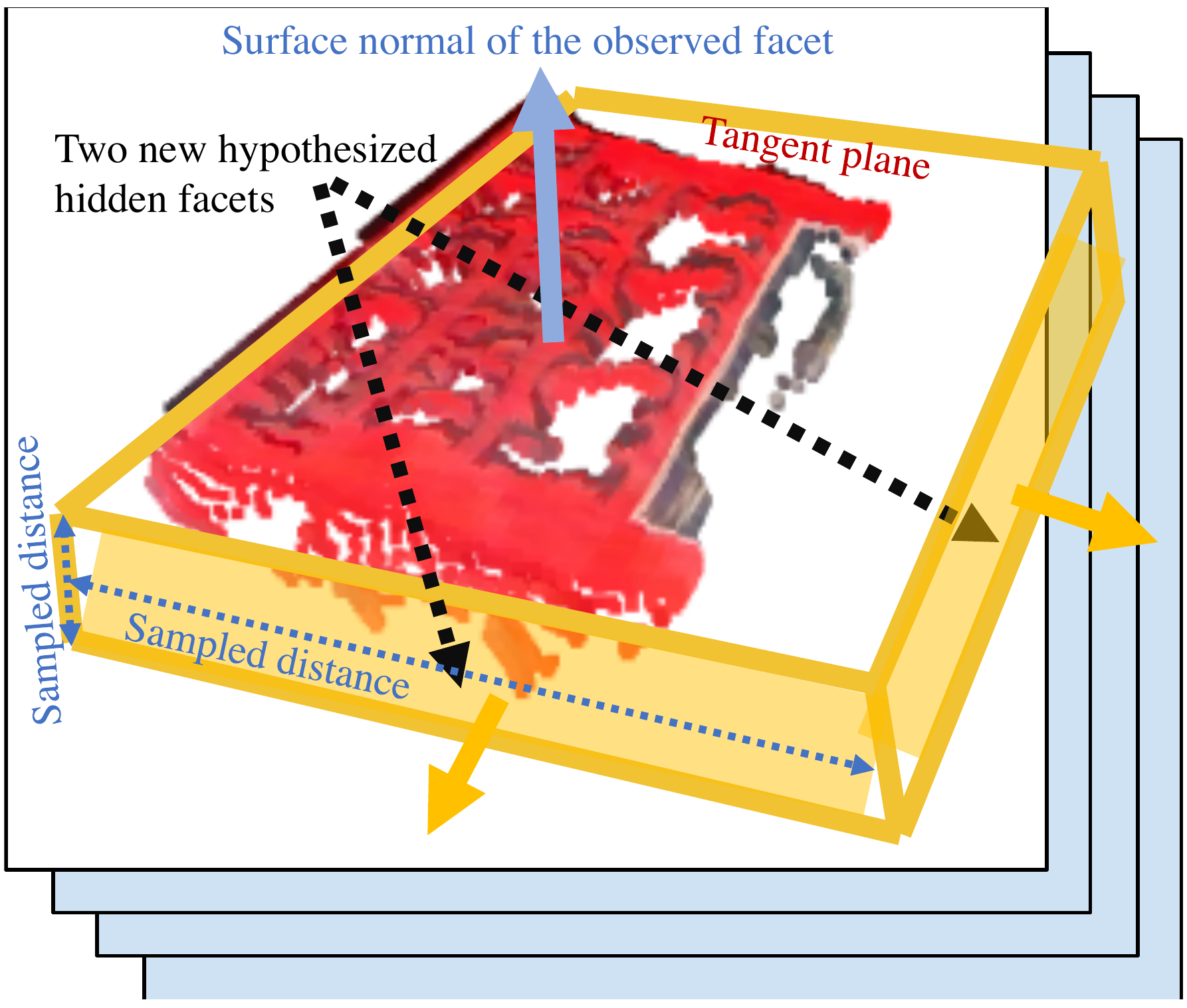}
\caption{\small {Sampling possible hidden facets of a partially occluded book from the scene of Figure~\ref{fig:pipeline}}}
\vspace{-0.4cm}
\label{fig:sampledSurfacesBook}
\end{wrapfigure}

Figure~\ref{fig:camera_ray} shows an example of a self-occluded object. The space occluded by the object defines the range of its hidden facets $\{F^h_j\}_{j=1}^{l}$. 
Any surface inside the invisible space could potentially belong to the object. Figure~\ref{fig:sampledSurfacesBook} shows an example of a hypothetical hidden surface of an object. Inferring hidden facets in the space of all possible 3D surfaces is computationally challenging for robotic manipulation tasks that require real-time inference. Therefore, we limit the space of hypotheses by exploiting the {\it Manhattan properties} that are commonly made in the literature~\cite{gao2016exploiting}. 
The Manhattan structure assumption states that the occluded facets have curvatures similar to the observed ones. 
This is not true in general but holds for most everyday objects.
Therefore, the first $m$ imagined facets are obtained by {\it mirroring} the $m$ observed facets along with their surface normals. Specifically, for each observed facet $F^o_j$ of an object we calculate the average surface normal of the facet and use the average tangent plane of the normal as a plane of symmetry. The point cloud of the observed facet $F^o_j$ is then mirrored along the tangent plane to generate a hypothesis facet $F^h_j$ after translating the mirrored facet along the opposite direction of the surface normal by a distance $d_j$. Distance $d_j$ is a {\it  with Monte Carlo Tree Searchfree parameter} that controls the position of $F^h_j$, it is iteratively sampled from an interval of $[D^{\textrm{min}}_j,D^{\textrm{max}}_j]$, where $D^{\textrm{min}}_j$ is
 the minimum length for objects to have a  volume, and $D^{\textrm{max}}_j$ is the maximum length. $D^{\textrm{max}}_j$, computed using {\it ray tracing}, ensures that no point in the space between the observed facet $F_j^o$ and its mirrored facet $F_j^h$ would belong to the visible volume of the scene. 

One would not be able to cover for all types of occlusions if  the hypothetical facets are limited to be $d_j$-distant mirror images of the observed facets, as described above. This solution covers only for self-occlusions. To account for occlusions caused by surrounding objects in clutter, we need to hypothesize additional facets. Consider the example of the book in Figure~\ref{fig:pipeline}. This book is inside a drawer and a significant part of it is occluded by the drawer's front. To solve these problems, we create a convex hull of all the facets (observed and hypothesized) every time we mirror the observed facets and we look for new facets in the convex hull.
The new facets are then inserted to the set $H_i$ that contains all hypothetical facets of object model $X_i$.
The new facets are also mirrored along their tangent planes, translated along new sampled distance, and inserted to set $H_i$.
This process is repeated until no new facets can be generated by mirroring or translating the existing ones without stepping out of the invisible space of the scene. A large number of models, with different volumes and geometries, can be generated with this procedure. The principal steps of this process are provided in Algorithm~\ref{sampling_algo}. 
Figure~\ref{fig:sampledSurfacesBook} shows how a hypothetical model of the object is sampled. We first mirror the only observed facet (part of the front cover) and translate it by a random distance. The convex hull of the two facets (front cover and hypothesized back cover) gives rise to six new side facets, which are also added to the set and mirrored in their turn to get different shapes and sizes of the book. This simple process, when repeated, can generate increasingly complex shapes.

\begin{algorithm}
\footnotesize
    \KwIn{A partial object model $O_i$ made of observed facets\;} 
    \KwOut{Set $H_i$ of hypothetiscal facets of object $O_i$\;}
    $H_i \leftarrow \emptyset $;$S \leftarrow O_i $\;
\Repeat{ $S = \emptyset$ or Timeout}{
\ForEach {$F \in S $}
{
Calculate $(\vec{N}, L)$, the average surface normal and the tangent plane at the center of facet $F$\;
Generate $F'$, the point cloud that is symmetrical to $F$ with respect to plane $L$\; 
Sample distance $x\sim \textrm{Uniform}(D_j^{\min},D_j^{\max})$\;
Translate each point in $F'$ by $-x\vec{N}$\;
$H_i \leftarrow H_i \cup \{F'\}$\;
}
Find $U$, the set of all facets in the convex hull of $H_i \cup O_i$\;
$S \leftarrow U - \Big(H_i \cup O_i\Big) $\;
Remove from $S$ all the facets that share the same surface normals as the ones already in $\Big(H_i \cup O_i\Big) $\; 
}
\caption{Hypothesis Generation}
\label{sampling_algo}
\end{algorithm}

\subsection{Global Geometric Constraints}
After performing the segmentation and facet decomposition steps described in Section~\ref{sec:seg_and_tracking}, 
we call Algorithm~\ref{sampling_algo} several times to sample a large number of different models for every detected object $i$. 
Each model $j$ of an object $i$ is a set $X^j_i = O_i\cup H^j_i$ made of observed facets set $O_i$, and generated facets set $H^j_i$. 
If the number of detected objects is $n$, and the number of  models per object is $m$, 
then the total set of hypotheses is  $\{X^j_i\}_{i=1}^n\textrm{\phantom .}_{j=1}^m$. 
In cluttered scenes, it is important to reason about combinations of models. 
What could look like a good model for an object may limit the choices of a neighboring object to unrealistic models.
Therefore, the generated hypotheses should satisfy certain geometric constraints, such that an object's surface cannot penetrate another object or the support surface, and a hypothesized hidden facet cannot intersect with the observed and known space of the scene.

We define a joint model for $n$ objects in the scene as an $n$-tuple $X = (X_1^{j_1} , X_2^{j_2}, \dots, X_n^{j_n})$. 
$\textrm{Constraints}(X,\{V_t\}_{t=0}^T)$ is a Boolean-valued function, defined as true if and only if:
\begin{align*}
\forall F , F' \in \cup_{i=1}^{n} X_{i}^{j_i}: (F  \neq F') \implies ( F  \cap  F' = \emptyset).
\end{align*}
The constraint implies that all the facets are distinct, which ensures that there are no nonempty intersections of objects. These geometric constraints immediately prune a large number of  hypotheses before starting the physics-based inference. 

\subsection{Inference Problem}
\label{sec:inference}
Given a sequence $\{\mu_t\}_{t=0}^{T}$ of pushing forces applied by the robot on the 3D points in the clutter along with the gravitational and normal forces, 
and a list ${\{O_{i,t}\}_{i=1}^n}_{t=0}^T$ of extracted partial models of $n$ objects obtained from segmentation, the problem consists in calculating
\begin{flalign}
\nonumber
&P( X | {\{O_{i,t}\}_{i=1}^n}_{t=0}^T, \{\mu_t\}_{t=0}^{T})\\ 
&\propto P({\{O_{i,t}\}_{i=1}^n}_{t=0}^T | X,\{\mu_t\}_{t=0}^{T} ) P(X),
\label{eq:bayes}
\end{flalign}
wherein 
$P(X)$ is a \textit{prior} of object models, which is uniform if the objects are completely unknown or a more informed distribution if the robot had already observed or manipulated similar objects, and 
$P({\{O_i\}_{i=1}^n}_{t=0}^T | X,\{\mu_t\}_{t=0}^{T} )$
is the \textit{likelihood} of the observations given a joint model $X$, which is described in the next section.
Note that $P(X) = 0$ for any model $X$ for which $\textrm{Constraints}(X,\{V_t\}_{t=0}^T)=\textrm{false}$. 
\begin{figure}[t]
\centering
\includegraphics[width=4.25cm,height=3.4cm]{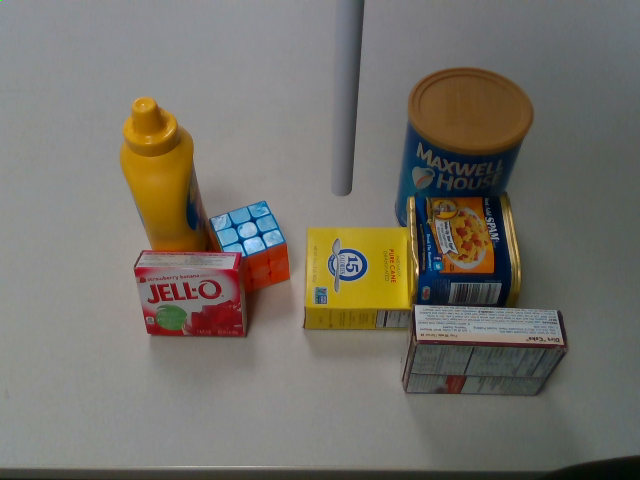}
\hspace{-0.3cm}
\includegraphics[trim={10cm 6cm 10cm 6cm},clip,width=4.25cm,height=3.4cm]{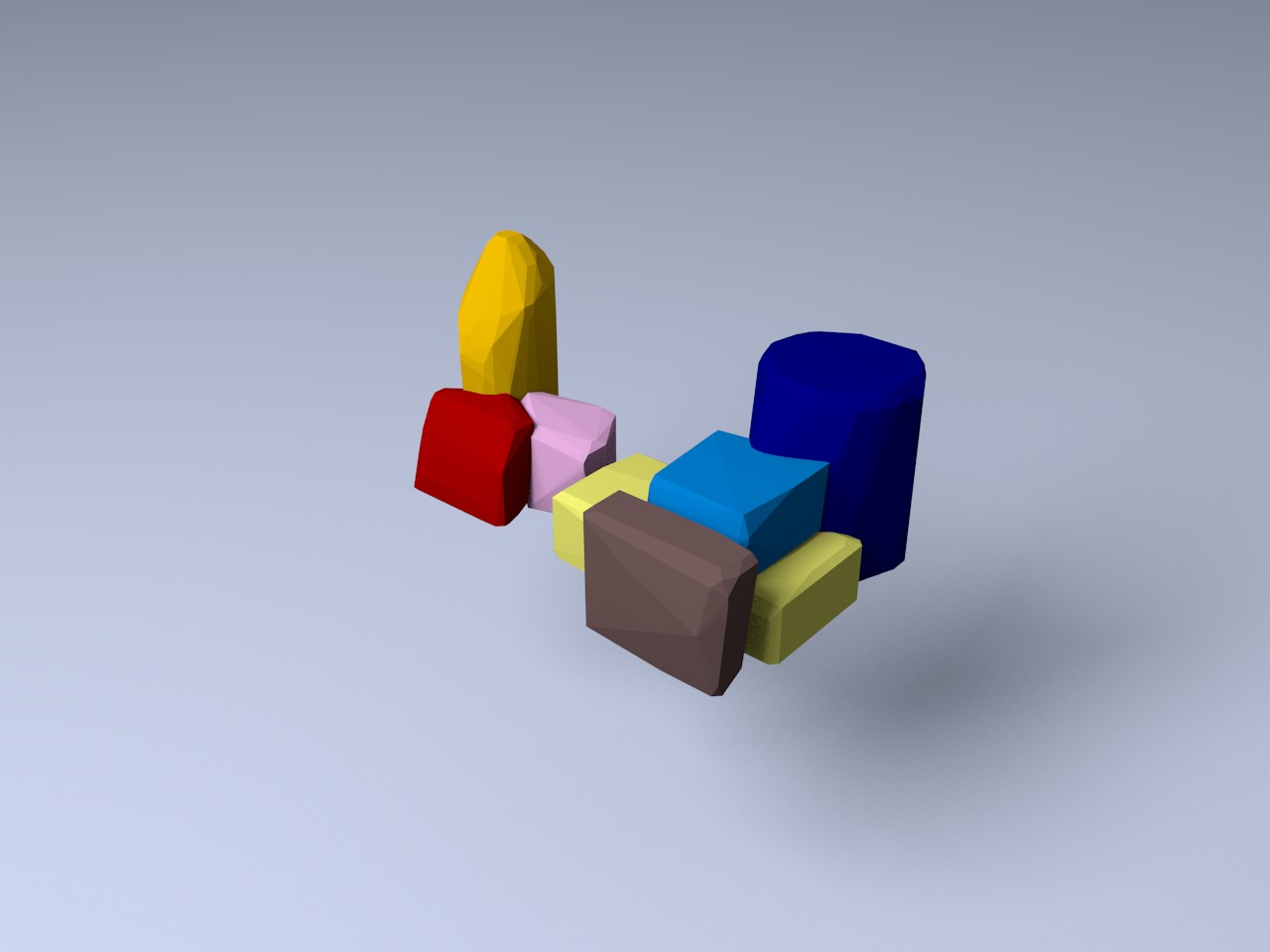}\\
\includegraphics[trim={10cm 6cm 10cm 8cm},clip,width=4.25cm,height=3.4cm]{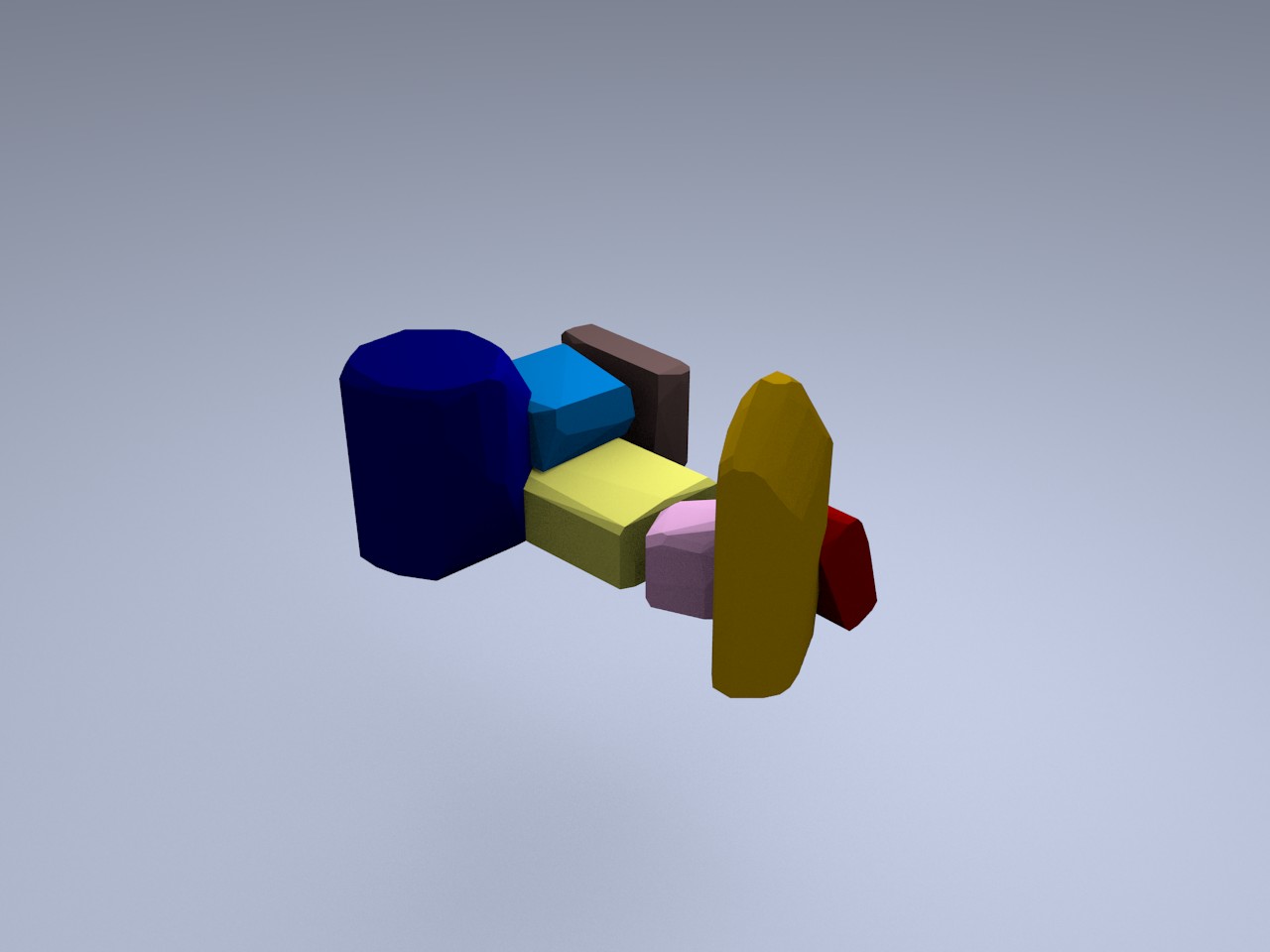}
\hspace{-0.3cm}
\includegraphics[trim={0 0 5.9cm 0},clip,width=4.25cm,height=3.4cm]{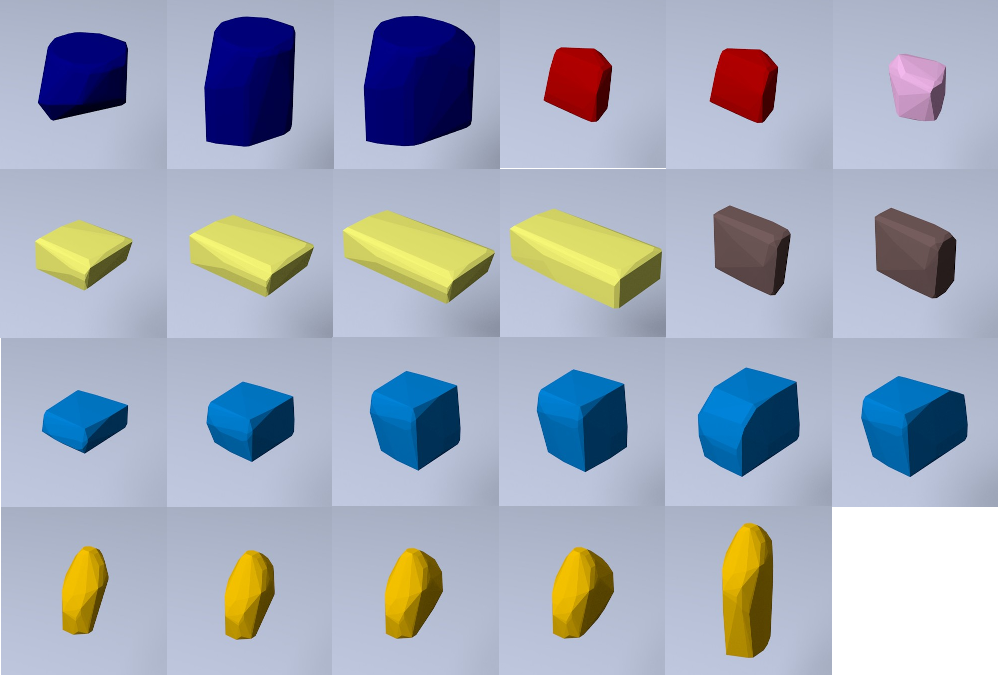}
\caption{An example of hypothesized shapes and reconstructed scene}
\vspace{-0.40cm}
\end{figure}

\subsection{Physical Likelihood Model}
\begin{figure}[h]
\centering
\begin{tabular}{cc}
\includegraphics[height=1.3cm,trim={0 0 0.5cm 0}]{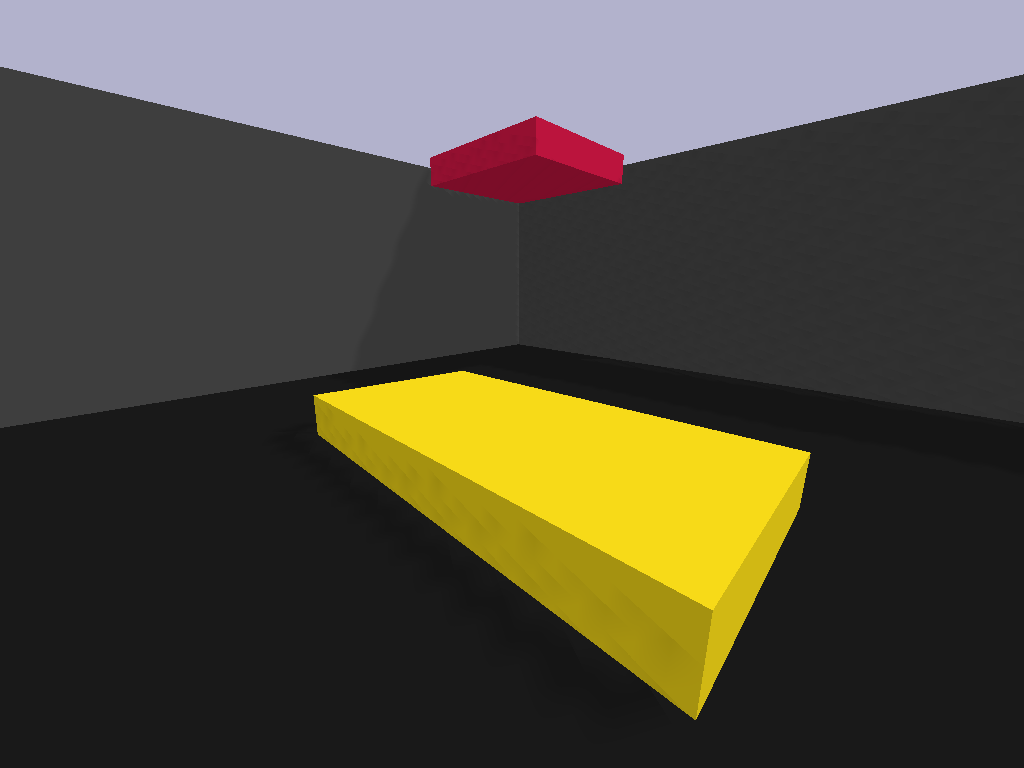} {\footnotesize $\rightarrow$}
\includegraphics[height=1.3cm,trim={0.5cm 0 0 0}]{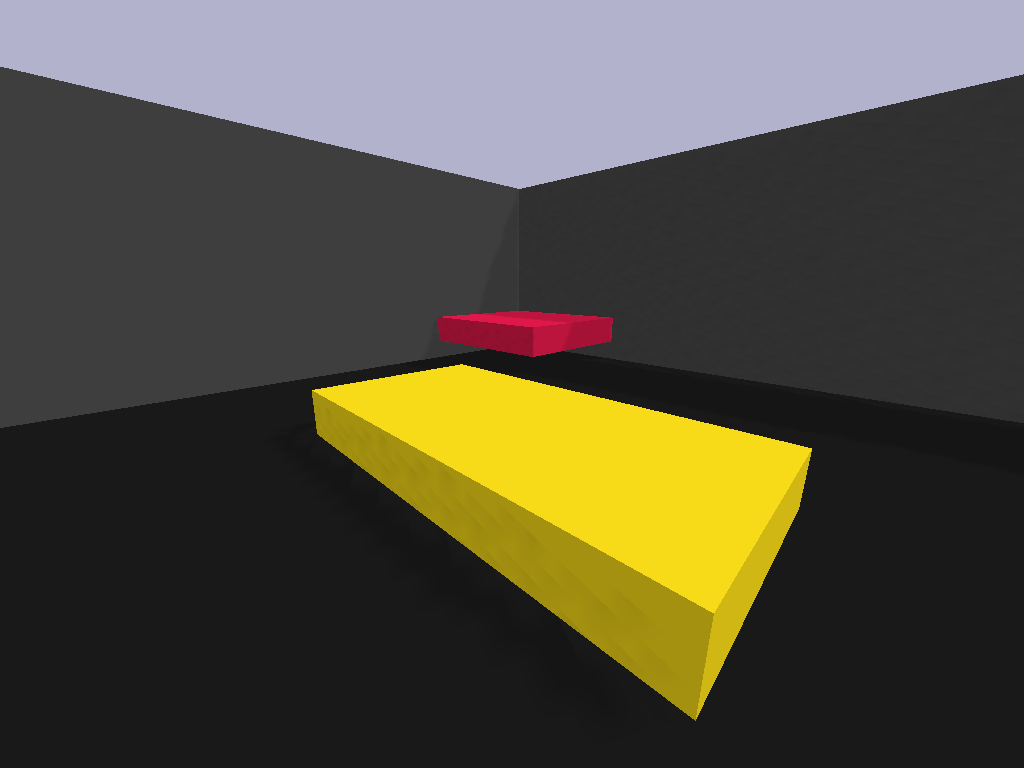}&
\includegraphics[height=1.3cm]{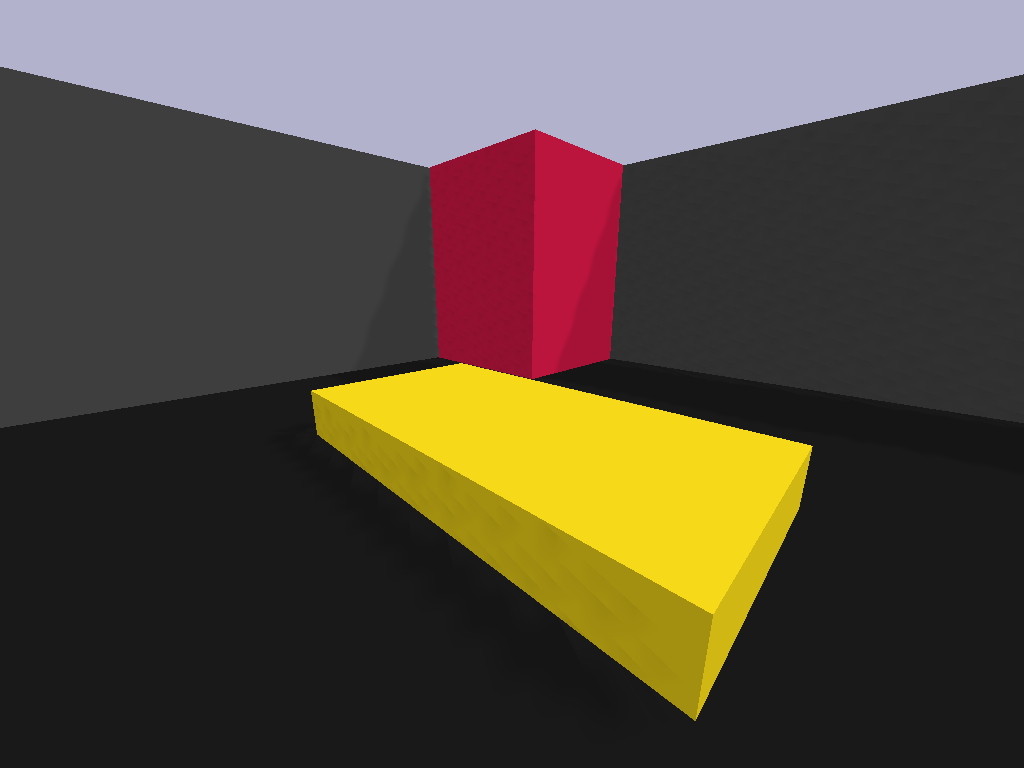}
{\footnotesize $\rightarrow$}
\includegraphics[height=1.3cm]{./img/simulation/sim0_gt.png}\\
Hypothesis (a) & Hypothesis (b)
\end{tabular}
\caption{\footnotesize Simulating the red box from the scene in Figures~\ref{fig:kuka} and~\ref{fig:pipeline}. The bottom of the box is occluded by the drawer. 
The top of the box falls down due to gravity in model (a) while it stands stable in (b) where the bottom part is hypothesized, which increases the probability of hypothesis (b).}
\label{fig:simulation_gravity}
\end{figure}

\begin{figure}[h]
\centering
\begin{tabular}{c c}
\includegraphics[height=1.3cm]{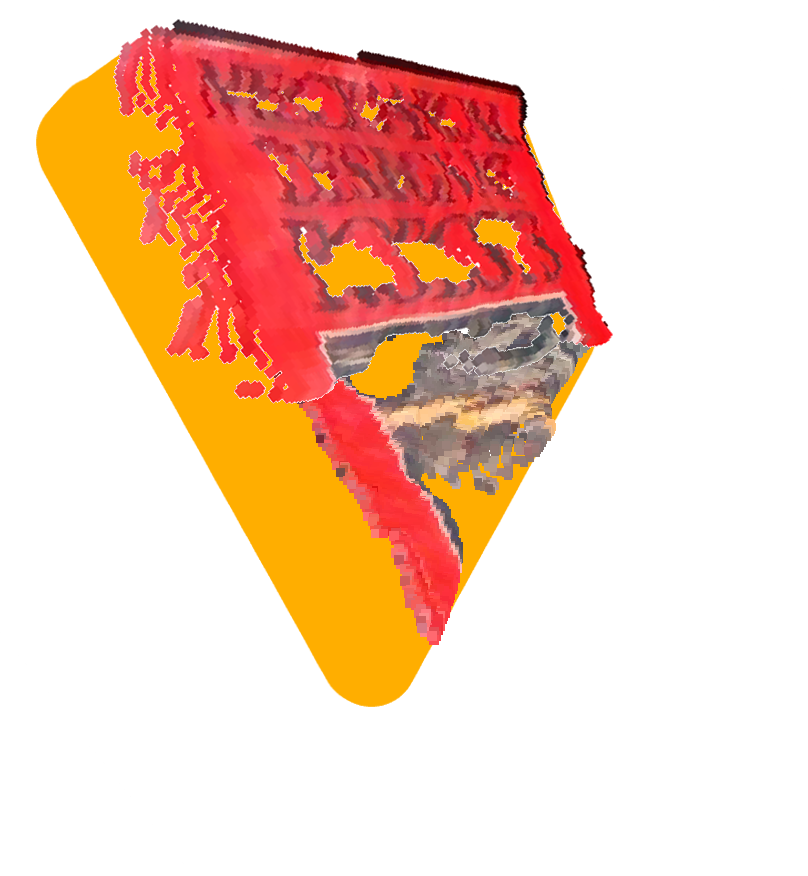}&
\includegraphics[height=1.4cm,trim={0 0 0.5cm 0}]{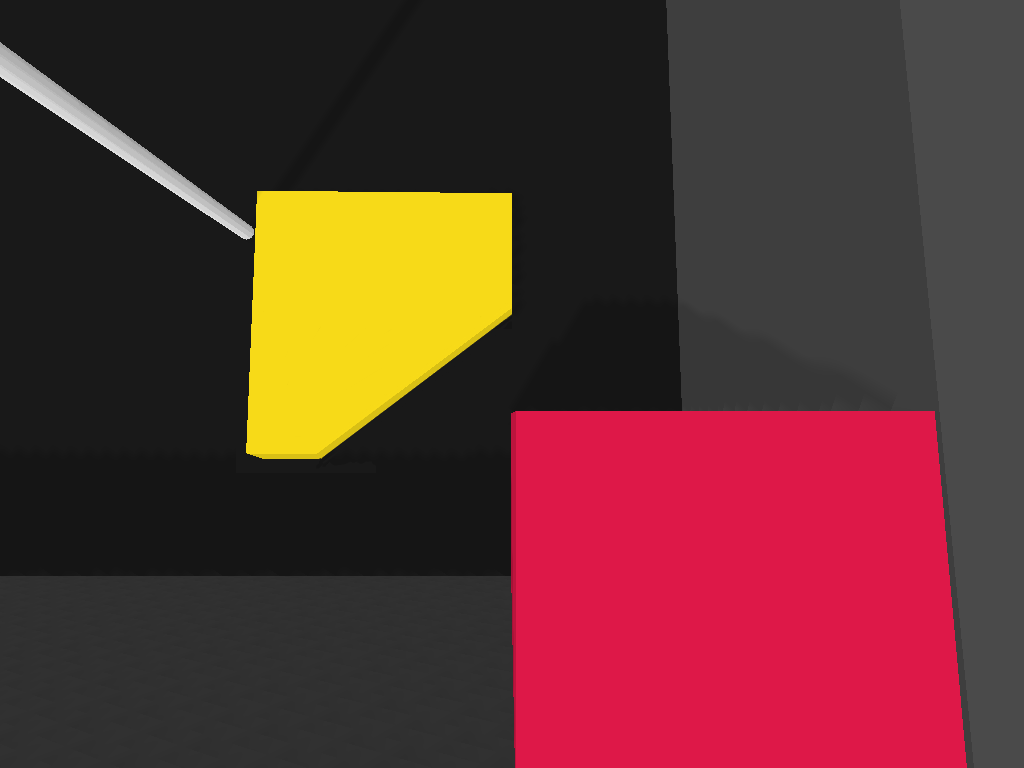}
\includegraphics[height=1.4cm,trim={0.5cm 0 0 0}]{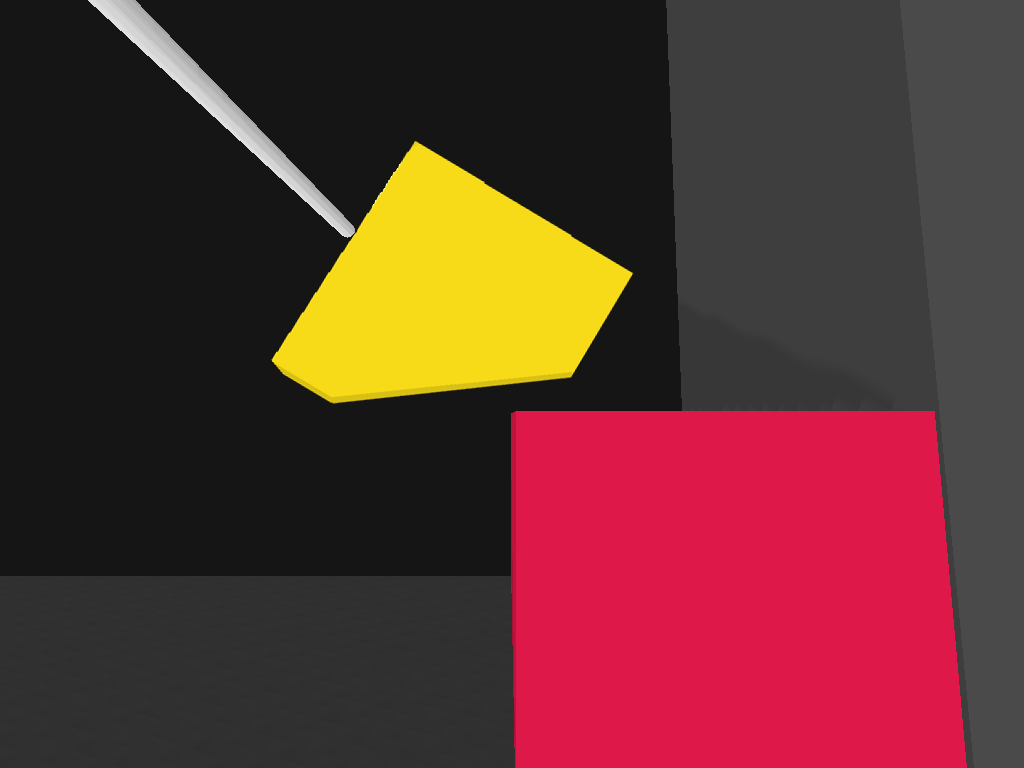}\\
\includegraphics[height=1.3cm]{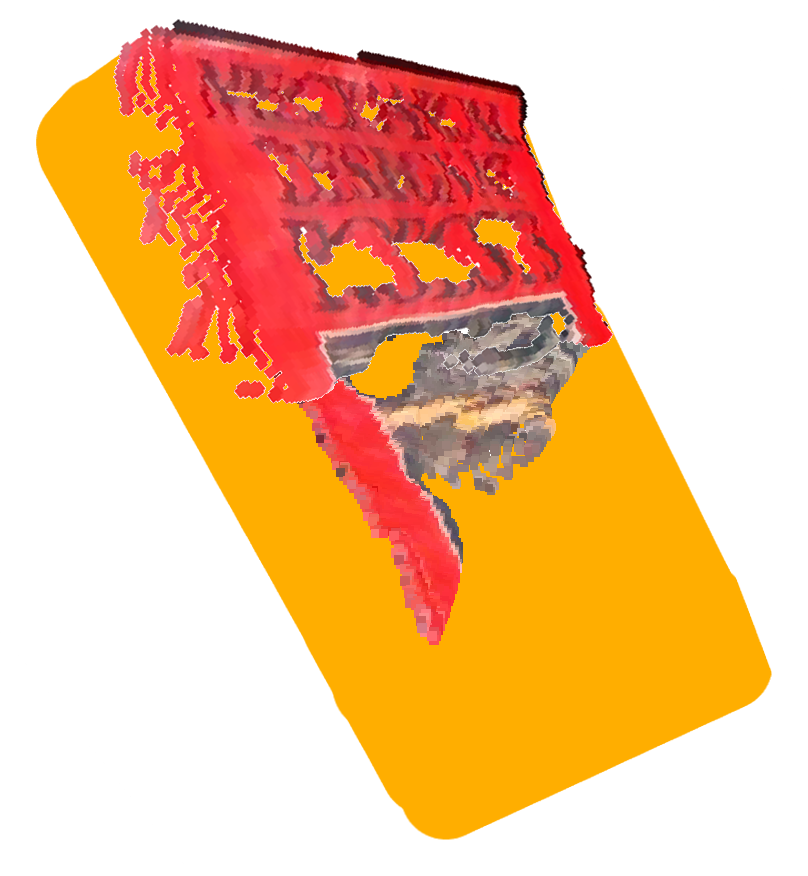}&
\includegraphics[height=1.4cm,trim={0 0 0.5cm 0}]{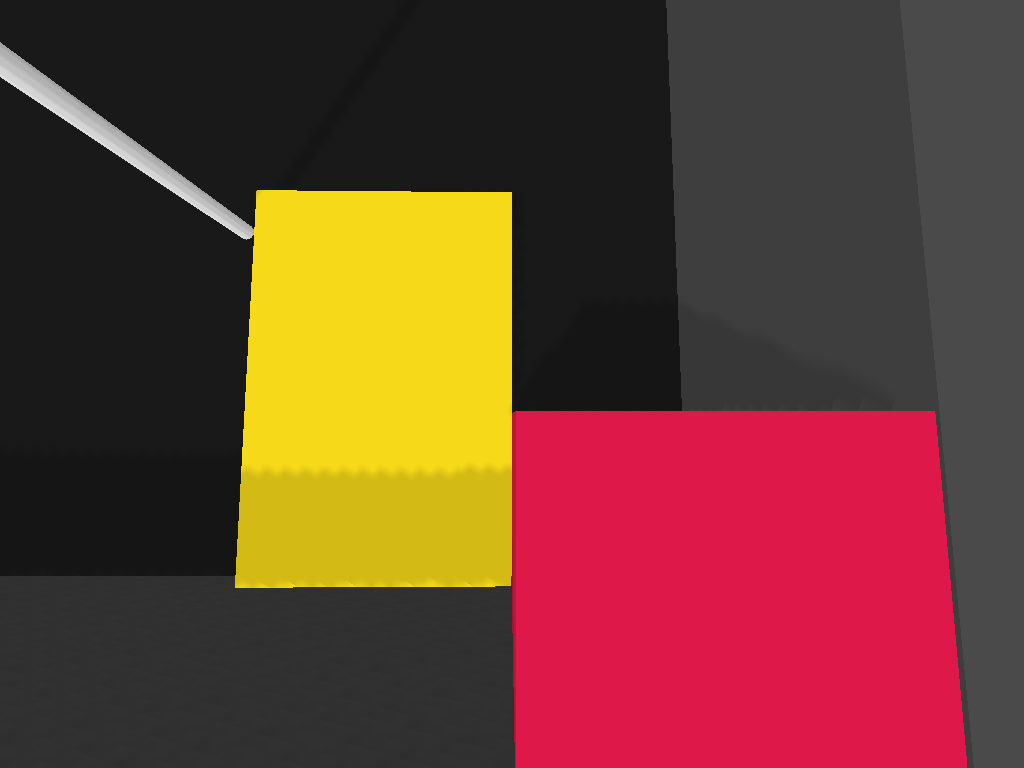}
\includegraphics[height=1.4cm,trim={0.5cm 0 0 0}]{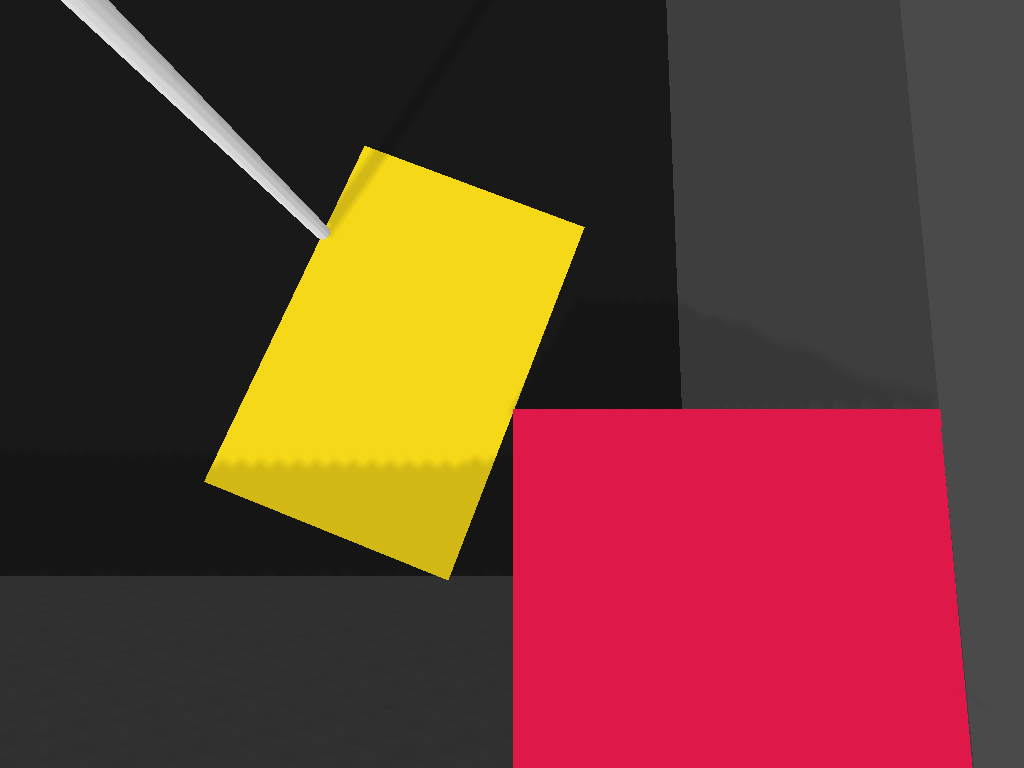}\\
\footnotesize  (a) Two sampled hypotheses & \footnotesize (b) Corresponding physics simulations\\
\end{tabular}
\caption{\footnotesize Inferring the shape of the book from the scene in Figures~\ref{fig:kuka} and~\ref{fig:pipeline}. The book, in yellow here, is adjacent to a red box. The white stick is the robot's end-effector pushing the book. The book is partially occluded by the drawer. 
Replaying the robot's horizontal pushing action in simulation using the bottom hypothesis predicts a rotation of the book that better matches with the real observation, compared to the small top model where the book moves more freely. Thus, the bottom hypothesis gets a higher probability. }
\vspace{-0.5cm}
\label{fig:simulation_action}
\end{figure}

We define likelihood $P({\{O_i\}_{i=1}^n}_{t=0}^T | X,\{\mu_t\}_{t=0}^{T} )$ as a function of the error between the current observation $O_t$ with pushing force $\mu_t$ and the  image predicted in simulation given object model $X$.
In other terms, the likelihood function quantifies the ability of a geometric model $X$ at predicting how the objects in the scene move under the effect of gravity and the robot's pushing actions. We take advantage of the availability of rigid-object simulators that can make such predictions.
In this work,
the \textit{Bullet}\footnote{http://bulletphysics.org} physics engine is utilized along with the \textit{Blender 3D renderer} for this purpose. The scene is recreated in simulation using each hypothesized joint model $X$. The objects are placed
in their initial positions by making sure that the observed facets have the same positions in simulation and in the initial real scene. 
All the forces exerted on the objects, including the robot's pokes and pushes as well as gravity, are simulated for time-steps $t\in\{0,\dots,T\}$. 
The likelihood function is then defined as
\begin{align}
&P(\{O_{i,t}\}_{i=1}^n\textrm{\phantom .}_{t=0}^T | X, \{\mu_t\}_{t=0}^{T} ) = \nonumber   \\ 
&\exp{ \Big( - \sum_{t=0}^{T} \sum_{i=1}^{n} \alpha \|  O_{i,t} - \hat{O}_i(X,\{\mu_k\}_{k=0}^{t} )  \|_2\Big)},
\label{eq:likelihood}
\end{align}
wherein $\hat{O}_i(X,\{\mu_k\}_{k=0}^{t})$ is the predicted depth image of object $i$ according to a given hypothesized joint model $X$ and given exerted forces $\{\mu_k\}_{k=0}^{t}$ up to time $t$.
This prediction is generated by rendering poses of all the objects.
The L2 distance is the difference between the observed depth image and the predicted one.
Note that the result depends on mechanical properties (friction and density), which are also unknown but can be searched along with the geometric model. We found out from our experiments that searching for friction and density is not necessary for the type of manipulation actions considered in this work. Thus, we use the same density and friction coefficient for all the objects in the simulation and we show in Section~\ref{sec:expr_sim} that the results are not sensitive to variations in density and friction. In fact, the forces applied by the robot on the objects are high enough to push them ahead but low enough to keep them in contact with the end effector. 
Figures~\ref{fig:simulation_gravity} and~\ref{fig:simulation_action} show intuitive examples of how the physical likelihood helps inferring more accurate shapes. 
\subsection{Inference through Monte-Carlo Tree Sampling}
Solving the inference problem of Section~\ref{sec:inference} is intractable in practice due to its combinatorial nature.
To compute $P(X|{\{O_{i,t}\}_{i=1}^n}_{t=0}^T, \{\mu_t\}_{t=0}^T)$, one needs to integrate the physics likelihood function over all possible hypothesized hidden facets of all objects, which has a complexity of $O(m^n)$ where $m$ is the number of model hypotheses and $n$ is the number of objects. 
Moreover, the integral of the marginal likelihood does not have a closed-form solution because of the discontinuities resulting from the collisions of the objects with each other.
We propose a Monte Carlo sampling method for approximating $P(X|{\{O_{i,t}\}_{i=1}^n}_{t=0}^T, \{\mu_t\}_{t=0}^T)$. This technique is explained in Algorithm~\ref{tree_algo}.

\begin{algorithm}[h!t]
\footnotesize
      \KwIn{Sequence of robotic actions  $\{\mu_t\}_{t=0}^T$, defined by their starting points, directions, and durations;
                Set ${\{O_{i,t}\}_{i=1}^n}_{t=0}^T$ of $n$ partial objects;
                Sequence $\{V_t\}_{t=0}^T$ of the visible spaces in the scene; Prior function $P(X)$, which is uniform by default.
               }
    \KwOut{Set of $m$ 3D models $\{X^j_i\}_{i=1}^n\textrm{\phantom .}_{j=1}^m$ for each one of the $n$ objects, and their estimated marginal posterior probabilities $\{P(X^j_i | \{(O_{i,t},\mu_t)\}_{t=0}^T )\}_{i=1}^n\textrm{\phantom .}_{j=1}^m$. }
  \tcc{\tiny Sample a large number of candidate shape models for each object}
  Use Algorithm~\ref{sampling_algo} to sample $m$ hypothetical models $\{X^j_i\}_{j=1}^m$ for each object $i\in {1,\dots, n}$, and 
  set $P(X^j_i)$ according to the prior\; 
\Repeat{ Timeout}{
  \tcc{\tiny Start with an empty scene, containing only support surfaces}
  \For{$(i := 0;\ i < n;\ i \leftarrow i + 1)$}{ placed [i] $\leftarrow$ false;
  model [i] $\leftarrow 0$ \; 
    \tcc{\tiny object $i$ has not yet been placed in the simulated scene}}

\For{$(stage := 1;\ stage \leq n;\ stage \leftarrow stage + 1)$}{
\tcc{\tiny Find an object to insert in the simulated scene}
	    max\textunderscore mass $\leftarrow$ 0\;
    	    \ForEach{ $i \in \{1,\dots,n\}$}{
	    mass $\leftarrow$ 0\;
	    \If{placed [i]  = true}{ continue \;}
    	    \ForEach{ $j \in \{1,\dots,m\}$}{
	    \tcc{\tiny Check if the scene remains stable after inserting object $i$ by using model $j$}
	    Create a scene with joint model $X$ wherein $i$ is placed using $X_i^{j}$, the objects that have been already placed in the previous stages are kept with their selected models, and the rest are placed using their minimum shapes\;
	    \If{( $\textrm{Constraints}(X,\{V_t\}_{t=0}^T) = $ false )}{ Exploration\textunderscore Prob[i,j] $\leftarrow$ $0$ \;}
	    \Else{
	    Simulate with joint model $X$ under gravity\;
	    Calculate $dist$, the distance by which object $i$ moved in the simulated scene\;
	    Exploration\textunderscore Prob[i,j] $\leftarrow$ $\exp(-\alpha dist)$ \;
	    mass $\leftarrow$ mass + Exploration\textunderscore Prob[i,j] \;
	    }
	    }
 	    \If{mass $\geq$ max\textunderscore mass}{max\textunderscore mass $\leftarrow$ mass; selected\textunderscore obj  $\leftarrow i$\;
	    	    \tcc{\tiny Select the object that causes the least disturbance when added to the scene}
	    }
	    \tcc{\tiny Normalize the exploration probabilities}
	    \ForEach{ $j \in \{1,\dots,m\}$ }{ \scriptsize   Exploration\textunderscore Prob[i,j] $\leftarrow$ Exploration\textunderscore Prob[i,j] /mass; }
	    }
		            j $\sim$ Exploration\textunderscore Prob[\textrm{selected\textunderscore obj},.];\tcc{\tiny sample a model}
		            model [\textrm{selected\textunderscore obj}] $\leftarrow j$\;
                    	    	           \tcc{\tiny Add the selected object to the scene}
		            placed [selected\textunderscore obj] $\leftarrow$ true \;
  }
 Create a complete initial scene with joint model $X$ wherein every object $i$ is assigned  to its sampled model $X^{model [i]}_i$ \;
 Simulate scene $X$ under gravity and robot's actions $\{\mu_t\}_{t=0}^T$\;
 	   \tcc{\tiny Compute likelihood with Equation~\ref{eq:likelihood} and update probabilities }
     	    \ForEach{ $i \in \{1,\dots,n\}$}{
             {\tiny $P(X^{model [i]}_i | \{O_{i,t}\}_{t=0}^T, \{\mu_t\}_{t=0}^T )
             \leftarrow P(\{O_{i,t}\}_{t=0}^T | X^{model [i]}_i,\{\mu_t\}_{t=0}^{T} )P(X^{model [i]}_i)/\textrm{Explor\textunderscore Prob[i,model[i]]}$}
             }
 }
  	   \tcc{\tiny Normalize the probabilities of the models for each object}
      	    \ForEach{ $i \in \{1,\dots,n\}$}{
	    	    mass = $\sum_{j=1}^m P(X^j_i | \{O_{i,t}\}_{t=0}^T, \{\mu_t\}_{t=0}^T )$;\tcp*{\tiny marginalization}
                      \ForEach{ $j \in \{1,\dots,m\}$}{ $P(X^j_i | \{O_{i,t}\}_{t=0}^T, \{\mu_t\}_{t=0}^T ) \leftarrow P(X^j_i | \{O_{i,t}\}_{t=0}^T, \{\mu_t\}_{t=0}^T ) / mass$; \tcp*{\tiny normalization} }
          	    }	    
\caption{\small{ Inverse Physics Reasoning (IPR) }}
\label{tree_algo}
\end{algorithm}

Algorithm~\ref{tree_algo} starts by generating a maximum number of candidate $3D$ models for each object (Line 1), by following the approach described in Algorithm~\ref{sampling_algo}.
The algorithm then tries to reconstruct, in a physics simulation, the initial scene before the robot's actions were executed (Lines 3-26). 
This reconstruction is performed by using a Monte Carlo Tree Search (MCTS) approach. 
Each attempt consists in placing the objects in the physics engine, one after another, according to the initial positions of their observed facets. 
At each stage, a new object is placed on top or next to the other objects in simulation, until the entire initial scene is reconstructed. 
Therefore, there is a set of $n-s+1$ objects left to choose from at a given stage $s$, these objects are indicated by the binary array $placed$. 
The order of placing the objects is important because objects that are on top of others cannot be placed before them. 
Moreover, each object $i$ has many candidate models $X_i^j$ that all match its observed facets. 
At each stage, we sample one model that we use for placing the selected object.
We use an exploration probability (Exploration\textunderscore Prob [i,j]) to sample a model $X_i^j$ for object $i$ (Lines 24-26).
Lines from 7 to 23 explain how the exploration probabilities are computed to focus the sampling on good models. The probability of using a model $X_i^j$ is proportional
to the stability of the scene that results from placing object $i$ with model $X_i^j$, 
while keeping the models of the already placed objects fixed, and using a minimum shape model for the other remaining objects.
The minimum shapes are made of only the observed facets. Subsequently, the object that is easiest to place 
(the one that can stand still on the support surface or on top of the already placed objects) is selected at each stage. 
At the end, the robot's actions are simulated on the fully reconstructed scene, 
and the probabilities of the sampled models are updated according to the similarity of the physics simulation to the actual observed motions of the facets in the real scene, using Equation~\ref{eq:likelihood}(Line 30).
\begin{wrapfigure}{l}{0.26\textwidth}
\centering
\vspace{-0.3cm}
\includegraphics[width=0.27\textwidth]{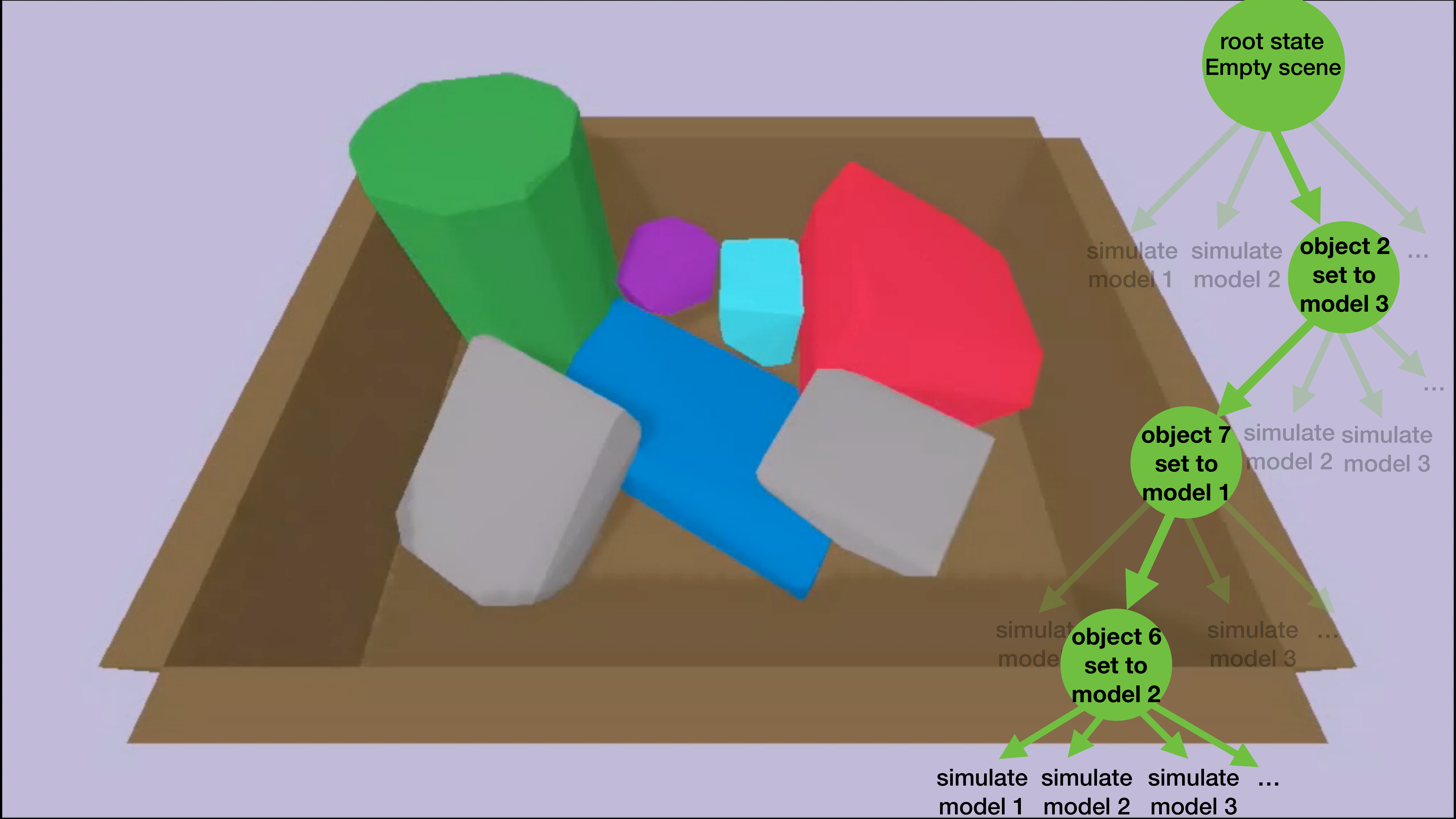}
\vspace{-0.4cm}
\caption{\footnotesize Scene reconstruction in a physics engine with Monte Carlo Tree Search} 
\label{fig:mcts}
\vspace{-0.5cm}
\end{wrapfigure}
Note that we also cancel out the sampling bias to ensure unbiased estimates by using {\it Importance Sampling}. 
This process is repeated all over, with different sampled models, until a timeout occurs. 
\vspace{0.1cm}
\section{Experiments}
We evaluated the proposed algorithm (IPR) in various scenes of unknown objects using the robotic platform in Figure~\ref{fig:kuka}. The corresponding datasets are described in Section~\ref{datasets}. We compared with recent alternative techniques, described in Section~\ref{Baselines}. The results are summarized in Section~\ref{results}.
\vspace{-0.5cm}
\subsection{Metrics}
 We report the average {\it Intersection over Union} (IoU) between the ground-truth occupied space of each object and its predicted occupied space. We also report the IoU between the entire occupied space of each scene and the union of the predicted 3D models of the objects within it, which is a weaker metric, but needed for some datasets (Voxlets).
\begin{figure}
\centering
\includegraphics[width=0.48\textwidth]{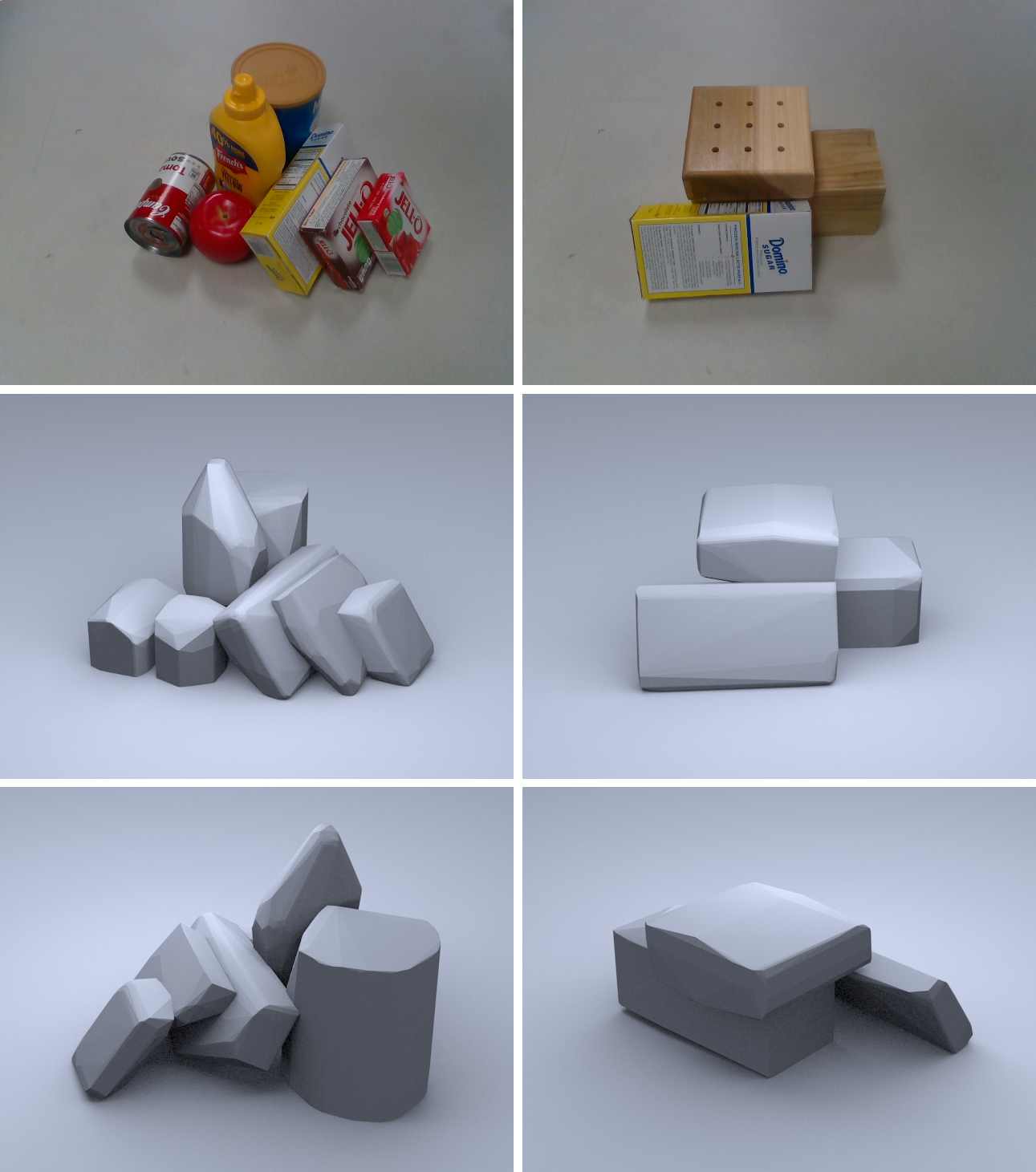}
\caption{\footnotesize Examples of our results on physics-based shape inference from a partial view; (top) input image of unknown objects; (middle and bottom) front and back views of the highest-probability hallucinated models.} 
\vspace{-0.5cm}
\label{fig:results_perception}
\end{figure}
\vspace{-0.1cm}
\subsection{Datasets}
\label{datasets}
Experiments are performed on two datasets: on a newly released \textit{\textit{Voxlets}} dataset\cite{firman-cvpr-2016}, and a dataset that we created using the {\it YCB benchmark}\cite{Callietal_RAM2015} objects.
The \textit{Voxlets} dataset contains static scenes of tabletop objects. 250 scenes are used for training and 30 are used for testing. This dataset does not contain ground-truth poses of individual objects, therefore we only evaluate the IoUs of entire scenes (union of objects). Our dataset with YCB objects includes the scenes shown in Figure~\ref{fig:results_perception} as well as piles of objects inside a tight box that can be seen in the attached video. This dataset is more challenging than
the \textit{\textit{Voxlets}} dataset because the piles are denser and contain more objects. Objects in this dataset are severely occluded. We split the dataset into two subsets, one with only static scenes and another with only dynamic ones. Static scenes are 12 in total. Dynamic scenes, 13 in total, include at least one robotic pushing action per scene.
We manually annotated the ground-truth voxel occupancy by fitting each object CAD model to the scenes.


\subsection{Methods}
\label{Baselines}
\textit{Zheng et. al.}\cite{Zheng_CVPR13} uses geometric and physics reasoning for recovering solid 3D volumetric primitives based on the Manhattan assumptions. This method, like ours, is {\it completely unsupervised} and well-suited for our setup. \textit{Voxlets}\cite{firman-cvpr-2016} is a learning-based method that predicts local geometry around observed points by employing a structured {\it Random Forest} classifier, which enables predicting shapes without any semantic understanding. It needs to be trained with a number of scenes, and it generalizes to new scenes. 
We trained \textit{Voxlets} with three different datasets: a) the original \textit{Voxlets} dataset~\cite{firman-cvpr-2016}, b) a synthetically generated YCB-object dataset of $10,000$ scenes, each containing 20 objects, and the objects in the scenes are different from the ones used in testing, and c), a synthetically generated YCB-object dataset of $10,000$ scenes that contains exactly the same objects and  angle of view that we used in the real testing scenes. 
\subsection{Variants of the Inverse Physics Reasoning (IPR)}
We performed an ablation study where we compare several variants of the IPR algorithm: 1) {\it Collision Checker} is IPR with a uniform prior on the object models minus the physics simulations, i.e. we only enforce the geometric constraints on the generated shapes. 2) {\it IPR+uniform} uses a uniform prior on the models of the objects, but simulates only gravity and collisions and does not simulate the robot's actions. 3) {\it IPR+size} is the same as the previous one, but uses a more informed prior where models with smaller volumes are given higher prior probabilities compared to large-sized models. 4) {\it IPR+action+uniform} is the same as {\it IPR+uniform} but also replays the robot's actions in simulation. 5) {\it IPR+action+size} is the same as {\it IPR+size} but also includes the robot's actions. 
\subsection{Results}
\label{results}
Table~\ref{tbl:voxlets} shows the results on the Voxlets dataset \cite{firman-cvpr-2016}. We followed the same evaluation metric as in \cite{firman-cvpr-2016}, where we calculate the IoU between piles instead of individual objects because the poses of objects in this dataset are missing. 
We did not compare to the variants of IPR with robotic actions because the scenes in Voxlets are all static. Both IPR+uniform and IPR+size achieved a higher IoU and recall than the other methods. Improvement over Collision Checker in particular shows that physics-based reasoning can help infer better models. Precision of IPR is comparable to other methods, but Zheng et. al. 2013~\cite{Zheng_CVPR13} has the highest precision because it predicts volume only where it is very certain, which makes the objects too small in general.
The Collision Checker has a performance that is very similar to  Zheng et. al. 2013~\cite{Zheng_CVPR13} because it is based on the same Manhattan assumptions and objects in the Voxlets dataset \cite{firman-cvpr-2016} are relatively away from each other. 
\begin{table}[h]
\centering
\begin{tabular}{|l|cccc|} \hline
Method & IoU & $F_1$ & prec. & recall\\ \hline
Zheng et. al. 2013~\cite{Zheng_CVPR13} & 0.571 & 0.729 & \textbf{0.839} & 0.645\\
Voxlets~\cite{firman-cvpr-2016} \scriptsize{(w/ Voxlets objects)}& 0.585 & 0.719 & 0.793 & 0.658 \\ \hline
Collision Checker  (ours)& 0.572 & 0.728 & 0.837 & 0.644 \\
IPR+uniform prior (ours)& 0.649 & 0.792 & 0.727 & 0.869\\
IPR+size prior (ours)& \textbf{0.663} & \textbf{0.803} & 0.768 & \textbf{0.841} \\ \hline
\end{tabular}
\caption{IoU on the \textit{Voxlets} dataset~\cite{firman-cvpr-2016}.}
\label{tbl:voxlets}
\vspace{-0.35cm}
\end{table}

Tables~\ref{tbl:ours_without_action} and~\ref{tbl:ours_with_action} show the results on our collected YCB dataset. Both tables are split into two parts: the bottom part is for the IoUs between each object and its predicted model, and the top part is for the IoU between each entire scene the union of all predicted models of objects in it. Table~\ref{tbl:ours_without_action} is for static scenes, while Table~\ref{tbl:ours_with_action} is for dynamic scenes where we can compare all variants of IPR. 
Results of per-object IoUs (bottom parts of the tables) are more relevant to robotics because it is important for motion planning and grasping to accurately infer shapes of individual objects.
IPR shows superior IoU in both sub-datasets as well as f-measure ($F_1=2\cdot\frac{\text{precision} \cdot \text{recall}} {\text{precision} + \text{recall}}$).
The physics simulation plays a major role in predicting the occluded volumes properly, as demonstrated by the fact that IPR outperforms its variant {\it Collision Checker} that  reasons only about geometries without including evidence from physics simulations of the scenes.

\begin{table}[h]
\centering
\begin{tabular}{|l| cccc |}
\multicolumn{5}{c}{Predicted scene space} \\\hline
Method & IoU & $F_1$ & prec. & recall\\ \hline
Zheng et. al. 2013~\cite{Zheng_CVPR13}                             & 0.485 & 0.654 & \textbf{0.887} & 0.518\\
Voxlets \cite{firman-cvpr-2016} \scriptsize{(w/ Voxlets objects)}  & 0.456 & 0.643 & 0.750 & 0.563\\
Voxlets \cite{firman-cvpr-2016} \scriptsize{(w/ diff. YCB objects)}& 0.416 & 0.604 & 0.618 & 0.590\\
Voxlets \cite{firman-cvpr-2016} \scriptsize{(w/ same YCB objects)} & 0.536 & 0.701 & 0.763 & 0.649 \\
\hline
Collision Checker & 0.485 & 0.654 & \textbf{0.887} & 0.518\\
IPR+uniform prior & 0.672 & 0.807 & 0.731 & \textbf{0.900}\\
IPR+size prior & \textbf{0.730} & \textbf{0.845} & 0.825 & 0.867 \\ \hline
\multicolumn{5}{l}{}\\
\multicolumn{5}{c}{Predicted object space} \\\hline
Method & IoU & $F_1$ & prec. & recall\\ \hline
Zheng et. al. 2013~\cite{Zheng_CVPR13}                             & 0.470 & 0.653 & \textbf{0.834} & 0.536\\
Voxlets \cite{firman-cvpr-2016} \scriptsize{(w/ Voxlets objects)}  & 0.411 & 0.604 & 0.469 & \textbf{0.849}\\
Voxlets \cite{firman-cvpr-2016} \scriptsize{(w/ diff. YCB objects)}& 0.476 & 0.675 & 0.569 & 0.829\\
Voxlets \cite{firman-cvpr-2016} \scriptsize{(w/ same YCB objects)} & 0.546 & 0.725 & 0.635 & 0.846 \\
\hline
Collision Checker       & 0.471 & 0.653 & \textbf{0.834} & 0.537\\
IPR+uniform prior       & 0.572 & 0.753 & 0.730 & 0.777 \\
IPR+size prior& \textbf{0.625} & \textbf{0.780}  & 0.790 & 0.771 \\ \hline
\end{tabular}
\caption{Average IoU in static scenes using YCB objects}
\label{tbl:ours_without_action}
\vspace{-0.35cm}
\end{table}

In Table~\ref{tbl:ours_with_action}, we can clearly see that replaying the robot's actions in simulation (IPR+action+uniform and IPR+action+size) significantly improves the IoU of objects. 
Unlike with the static scenes in Table~\ref{tbl:ours_without_action}, the size prior does not help a lot when the robot's actions are already taken into account in computing the likelihood of hypothesized models.

\begin{table}[h]
\centering
\begin{tabular}{|l| cccc |}
\multicolumn{5}{c}{Predicted scene space} \\\hline
Method & IoU & $F_1$ & prec. & recall\\ \hline
Zheng et. al. 2013~\cite{Zheng_CVPR13}                             & 0.501 & 0.667 & \textbf{0.897} & 0.538\\
Voxlets \cite{firman-cvpr-2016} \scriptsize{(w/ Voxlets objects)}  & 0.413 & 0.597 & 0.531 & 0.682\\
Voxlets \cite{firman-cvpr-2016} \scriptsize{(w/ diff. YCB objects)}& 0.388 & 0.559 & 0.473 & 0.683\\
Voxlets \cite{firman-cvpr-2016} \scriptsize{(w/ same YCB objects)} & 0.423 & 0.594 & 0.518 & 0.695 \\ 
\hline
Collision Checker & 0.499 & 0.667 & 0.882 & 0.536\\
IPR+uniform prior & 0.694 & 0.822 & 0.792 & \textbf{0.854} \\
IPR+action+uniform prior & \textbf{0.702} & \textbf{0.828} & 0.819 & 0.837 \\ 
IPR+action+size prior & 0.700 & 0.826 & 0.839 & 0.813 \\
\hline
\multicolumn{5}{l}{}\\
\multicolumn{5}{c}{Predicted object space} \\\hline
Method & IoU & $F_1$ & prec. & recall\\ \hline
Zheng et. al. 2013~\cite{Zheng_CVPR13}                             & 0.474 & 0.650 & 0.837 & 0.531\\
Voxlets \cite{firman-cvpr-2016} \scriptsize{(w/ Voxlets objects)}  & 0.370 & 0.551 & 0.412 & 0.831\\
Voxlets \cite{firman-cvpr-2016} \scriptsize{(w/ diff. YCB objects)}& 0.489 & 0.677 & 0.580 &	0.813\\
Voxlets \cite{firman-cvpr-2016} \scriptsize{(w/ same YCB objects)} & 0.516 & 0.692 & 0.589 & \textbf{0.839}\\ 
\hline
Collision Checker & 0.478 & 0.655 & \textbf{0.844} & 0.535\\
IPR+uniform prior & 0.618 & 0.777 &	0.773 &	0.782 \\
IPR+action+uniform prior & \textbf{0.640} & \textbf{0.793} & 0.795 & 0.792\\ 
IPR+action+size prior & 0.638 & 0.789 & 0.814 & 0.766 \\
\hline
\end{tabular}
\caption{Average IoU in dynamic scenes using YCB objects}
\label{tbl:ours_with_action}
\vspace{-0.35cm}
\end{table}

We measured the average computation time per object in the dynamic scenes: Zheng et. al. 2013~\cite{Zheng_CVPR13} took $0.34$ seconds, Voxlets~\cite{firman-cvpr-2016} took $21.71$ seconds, Collision Checker took $0.32$ seconds, and the full IPR (IPR + action + prior) method took $21.75$ seconds.
IPR takes a comparable computation time as Voxlets~\cite{firman-cvpr-2016} while it achieves a significantly higher accuracy.
The computation time of IPR with exhaustive search (instead of Monte Carlo) is $115.09$ seconds. 
The hypothesis generation step takes $7.75$ seconds per object. Full IPR has only $13.04\%$ of the exhaustive search's computational burden, if we exclude the hypothesis generation preprocessing step which is common to both methods.
\subsection{Physics Simulation with Unknown Mechanical Properties}
\label{sec:expr_sim}
The uncertainty regarding mechanical properties (friction and volumetric mass density) of objects can cause different simulation results even when the same object shape is used. To verify the real impact of these properties on our results, we sampled $1,000$ different values of mass densities and friction coefficients in the ranges between the maximum and minimum of mass density and friction values of the entire YCB objects dataset. The friction ranges were obtained from~\cite{engineering_toolbox}. We simulated the motions of the sampled mechanical models of objects under gravity and the robot's pushing actions and we found that the standard deviation of the objects' positions is $0.658cm$, which is negligible considering that we down-sampled the input point clouds into 3D voxels of $0.5cm$ and the noise in the point cloud is within the same order. This result holds only when the range of the mechanical properties of the objects is not too large. The general problem of inferring simultaneously 3D and mechanical models will be the subject of a future work. 

\small

\bibliographystyle{IEEEtran}
\bibliography{cite.bib}

\begin{thebibliography}{10}
\providecommand{\url}[1]{#1}
\csname url@rmstyle\endcsname
\providecommand{\newblock}{\relax}
\providecommand{\bibinfo}[2]{#2}
\providecommand\BIBentrySTDinterwordspacing{\spaceskip=0pt\relax}
\providecommand\BIBentryALTinterwordstretchfactor{4}
\providecommand\BIBentryALTinterwordspacing{\spaceskip=\fontdimen2\font plus
\BIBentryALTinterwordstretchfactor\fontdimen3\font minus
  \fontdimen4\font\relax}
\providecommand\BIBforeignlanguage[2]{{%
\expandafter\ifx\csname l@#1\endcsname\relax
\typeout{** WARNING: IEEEtran.bst: No hyphenation pattern has been}%
\typeout{** loaded for the language `#1'. Using the pattern for}%
\typeout{** the default language instead.}%
\else
\language=\csname l@#1\endcsname
\fi
#2}}

\bibitem{specialissuemanipulation2013}
H.~B. Amor, A.~Saxena, N.~Hudson, and J.~Peters, Eds., \emph{{Special Issue on
  Autonomous Grasping and Manipulation}}.\hskip 1em plus 0.5em minus
  0.4em\relax Springer: Autonomous Robots, 2013.

\bibitem{Bohg}
J.~Bohg, A.~Morales, T.~Asfour, and D.~Kragic, ``{Data-Driven Grasp Synthesis -
  A Survey},'' \emph{IEEE Transactions on Robotics}, pp. 289--309, 2013.

\bibitem{LaValle:2006:PA:1213331}
S.~M. LaValle, \emph{Planning Algorithms}.\hskip 1em plus 0.5em minus
  0.4em\relax New York, NY, USA: Cambridge University Press, 2006.

\bibitem{7583659}
N.~Correll, K.~E. Bekris, D.~Berenson, O.~Brock, A.~Causo, K.~Hauser, K.~Okada,
  A.~Rodriguez, J.~M. Romano, and P.~R. Wurman, ``Analysis and observations
  from the first amazon picking challenge,'' \emph{IEEE Transactions on
  Automation Science and Engineering}, vol.~15, no.~1, pp. 172--188, Jan 2018.

\bibitem{RennieSBS16}
C.~Rennie, R.~Shome, K.~E. Bekris, and A.~F.~D. Souza, ``A dataset for improved
  rgbd-based object detection and pose estimation for warehouse
  pick-and-place,'' \emph{{IEEE} Robotics and Automation Letters}, vol.~1,
  no.~2, pp. 1179--1185, 2016.

\bibitem{Callietal_RAM2015}
B.~Calli, A.~Walsman, A.~Singh, S.~Srinivasa, P.~Abbeel, and A.~Dollar,
  ``Benchmarking in manipulation research: The ycb object and model set and
  benchmarking protocols,'' \emph{IEEE Robotics and Automation Magazine (RAM)},
  2015.

\bibitem{Janoch2011}
\emph{A Category-Level 3-D Object Dataset: Putting the Kinect to Work},
  November 2011.

\bibitem{furrer2017}
F.~Furrer, M.~Wermelinger, H.~Yoshida, F.~Gramazio, M.~Kohler, R.~Siegwart, and
  M.~Hutter, ``Autonomous robotic stone stacking with online next best object
  target pose planning,'' in \emph{2017 IEEE International Conference on
  Robotics and Automation (ICRA)}, May 2017, pp. 2350--2356.

\bibitem{6696928}
C.~Eppner and O.~Brock, ``Grasping unknown objects by exploiting shape
  adaptability and environmental constraints,'' in \emph{2013 IEEE/RSJ
  International Conference on Intelligent Robots and Systems}, Nov 2013, pp.
  4000--4006.

\bibitem{SungLS17}
J.~Sung, I.~Lenz, and A.~Saxena, ``Deep multimodal embedding: Manipulating
  novel objects with point-clouds, language and trajectories,'' in
  \emph{{ICRA}}.\hskip 1em plus 0.5em minus 0.4em\relax {IEEE}, 2017, pp.
  2794--2801.

\bibitem{PintoG16}
L.~Pinto and A.~Gupta, ``Supersizing self-supervision: Learning to grasp from
  50k tries and 700 robot hours,'' in \emph{{ICRA}}.\hskip 1em plus 0.5em minus
  0.4em\relax {IEEE}, 2016, pp. 3406--3413.

\bibitem{KroemerJRAS_66360}
O.~Kroemer, R.~Detry, J.~Piater, and J.~Peters, ``Combining active learning and
  reactive control for robot grasping,'' no.~9, pp. 1105--1116, 2010.

\bibitem{detry2017c}
R.~Detry, J.~Papon, and L.~Matthies, ``Task-oriented grasping with semantic and
  geometric scene understanding,'' in \emph{{IEEE/RSJ} International Conference
  on Intelligent Robots and Systems}, 2017.

\bibitem{4543223}
G.~M. Bone, A.~Lambert, and M.~Edwards, ``Automated modeling and robotic
  grasping of unknown three-dimensional objects,'' in \emph{2008 IEEE
  International Conference on Robotics and Automation}, May 2008, pp. 292--298.

\bibitem{Bouaziz:2013}
S.~Bouaziz, A.~Tagliasacchi, and M.~Pauly, ``Sparse iterative closest point,''
  in \emph{Proceedings of the Eleventh Eurographics/ACMSIGGRAPH Symposium on
  Geometry Processing}, ser. SGP '13.\hskip 1em plus 0.5em minus 0.4em\relax
  Aire-la-Ville, Switzerland, Switzerland: Eurographics Association, 2013, pp.
  113--123.

\bibitem{fusionpp2018}
J.~McCormac*, R.~Clark*, M.~Bloesch, A.~J. Davison, and S.~Leutenegger,
  ``Fusion++: Volumetric object-level slam,'' \emph{International Conference on
  3DVision}, 2018, * Joint first authors.

\bibitem{KahnSPBRGA15}
G.~Kahn, P.~Sujan, S.~Patil, S.~Bopardikar, J.~Ryde, K.~Y. Goldberg, and
  P.~Abbeel, ``Active exploration using trajectory optimization for robotic
  grasping in the presence of occlusions,'' in \emph{{IEEE} International
  Conference on Robotics and Automation, {ICRA} 2015, Seattle, WA, USA, 26-30
  May, 2015}, 2015, pp. 4783--4790.

\bibitem{KraininCF11}
M.~Krainin, B.~Curless, and D.~Fox, ``Autonomous generation of complete 3d
  object models using next best view manipulation planning,'' in \emph{{IEEE}
  International Conference on Robotics and Automation, {ICRA} 2011, Shanghai,
  China, 9-13 May 2011}, 2011, pp. 5031--5037.

\bibitem{LPKTLPICRA12}
L.~P. Kaelbling and T.~Lozano-Perez, ``Unifying perception, estimation and
  action for mobile manipulation via belief space planning,'' in \emph{IEEE
  Conference on Robotics and Automation (ICRA)}, 2012.

\bibitem{6631288}
M.~R. Dogar, M.~C. Koval, A.~Tallavajhula, and S.~S. Srinivasa, ``Object search
  by manipulation,'' in \emph{2013 IEEE International Conference on Robotics
  and Automation}, May 2013, pp. 4973--4980.

\bibitem{8007233}
J.~Bohg, K.~Hausman, B.~Sankaran, O.~Brock, D.~Kragic, S.~Schaal, and G.~S.
  Sukhatme, ``Interactive perception: Leveraging action in perception and
  perception in action,'' \emph{IEEE Transactions on Robotics}, vol.~33, no.~6,
  pp. 1273--1291, Dec 2017.

\bibitem{4543220}
D.~Katz and O.~Brock, ``Manipulating articulated objects with interactive
  perception,'' in \emph{2008 IEEE International Conference on Robotics and
  Automation}, May 2008, pp. 272--277.

\bibitem{hoof2014probabilistic}
H.~van Hoof, O.~Kroemer, and J.~Peters, ``Probabilistic segmentation and
  targeted exploration of objects in cluttered environments,'' no.~5, pp.
  1198--1209, 2014.

\bibitem{7298863}
J.~Rock, T.~Gupta, J.~Thorsen, J.~Gwak, D.~Shin, and D.~Hoiem, ``Completing 3d
  object shape from one depth image,'' in \emph{2015 IEEE Conference on
  Computer Vision and Pattern Recognition (CVPR)}, 2015, pp. 2484--2493.

\bibitem{DaiQN17}
A.~Dai, C.~R. Qi, and M.~Nie{\ss}ner, ``Shape completion using
  3d-encoder-predictor cnns and shape synthesis,'' in \emph{2017 {IEEE}
  Conference on Computer Vision and Pattern Recognition, {CVPR} 2017, Honolulu,
  HI, USA, July 21-26, 2017}, 2017, pp. 6545--6554.

\bibitem{VarleyDRRA17}
J.~Varley, C.~DeChant, A.~Richardson, J.~Ruales, and P.~K. Allen, ``Shape
  completion enabled robotic grasping,'' in \emph{2017 {IEEE/RSJ} International
  Conference on Intelligent Robots and Systems, {IROS} 2017, Vancouver, BC,
  Canada, September 24-28, 2017}, 2017, pp. 2442--2447.

\bibitem{WuSKYZTX15}
Z.~Wu, S.~Song, A.~Khosla, F.~Yu, L.~Zhang, X.~Tang, and J.~Xiao, ``3d
  shapenets: A deep representation for volumetric shapes.'' in
  \emph{CVPR}.\hskip 1em plus 0.5em minus 0.4em\relax IEEE Computer Society,
  2015, pp. 1912--1920.

\bibitem{jrock-cvpr-2015}
J.~Rock, T.~Gupta, J.~Thorsen, J.~Gwak, D.~Shin, and D.~Hoiem, ``Completing 3d
  object shape from one depth image,'' in \emph{2015 IEEE Conference on
  Computer Vision and Pattern Recognition (CVPR)}, June 2015, pp. 2484--2493.

\bibitem{firman-cvpr-2016}
M.~Firman, O.~M. Aodha, S.~Julier, and G.~J. Brostow, ``{Structured Completion
  of Unobserved Voxels from a Single Depth Image},'' in \emph{Computer Vision
  and Pattern Recognition (CVPR)}, 2016.

\bibitem{dai2017complete}
A.~Dai, C.~R. Qi, and M.~Nie{\ss}ner, ``Shape completion using
  3d-encoder-predictor cnns and shape synthesis,'' in \emph{Proc. Computer
  Vision and Pattern Recognition (CVPR), IEEE}, 2017.

\bibitem{Kimia2003}
B.~B. Kimia, I.~Frankel, and A.-M. Popescu, ``Euler spiral for shape
  completion,'' \emph{International Journal of Computer Vision}, vol.~54,
  no.~1, pp. 159--182, Aug 2003.

\bibitem{Attene2006}
M.~Attene, B.~Falcidieno, and M.~Spagnuolo, ``Hierarchical mesh segmentation
  based on fitting primitives,'' \emph{The Visual Computer}, vol.~22, no.~3,
  pp. 181--193, Mar 2006.

\bibitem{gao2016exploiting}
Y.~Gao and A.~L. Yuille, ``Exploiting symmetry and/or manhattan properties for
  3d object structure estimation from single and multiple images,'' in
  \emph{IEEE International Conference on Computer Vision and Pattern
  Recognition}, 2017.

\bibitem{Zheng_CVPR13}
B.~Zheng, Y.~Zhao, J.~C. Yu, K.~Ikeuchi, and S.-C. Zhu, ``Beyond point clouds:
  Scene understanding by reasoning geometry and physics.'' in
  \emph{CVPR}.\hskip 1em plus 0.5em minus 0.4em\relax IEEE Computer Society,
  2013, pp. 3127--3134.

\bibitem{smzkxzm_imaginingTheUnseen_sigga14}
T.~Shao*, A.~Monszpart*, Y.~Zheng, B.~Koo, W.~Xu, K.~Zhou, and N.~Mitra,
  ``Imagining the unseen: Stability-based cuboid arrangements for scene
  understanding,'' \emph{ACM SIGGRAPH Asia 2014}, 2014, * Joint first authors.

\bibitem{DBLP:conf/cogsci/HamrickBT11}
J.~B. Hamrick, P.~Battaglia, and J.~B. Tenenbaum, ``Probabilistic internal
  physics models guide judgments about object dynamics,'' in
  \emph{CogSci}.\hskip 1em plus 0.5em minus 0.4em\relax
  cognitivesciencesociety.org, 2011.

\bibitem{Battaglia18327}
P.~W. Battaglia, J.~B. Hamrick, and J.~B. Tenenbaum, ``Simulation as an engine
  of physical scene understanding,'' \emph{Proceedings of the National Academy
  of Sciences}, vol. 110, no.~45, pp. 18\,327--18\,332, 2013.

\bibitem{0003LF17}
W.~Li, A.~Leonardis, and M.~Fritz, ``Visual stability prediction for robotic
  manipulation,'' in \emph{2017 {IEEE} International Conference on Robotics and
  Automation, {ICRA}}, 2017, pp. 2606--2613.

\bibitem{Stein2014}
S.~C. Stein, M.~Schoeler, J.~Papon, and F.~W\"{o}rg\"{o}tter, ``Object
  partitioning using local convexity,'' in \emph{Proceedings of the 2014 IEEE
  Conference on Computer Vision and Pattern Recognition}, ser. CVPR '14.\hskip
  1em plus 0.5em minus 0.4em\relax Washington, DC, USA: IEEE Computer Society,
  2014, pp. 304--311.

\bibitem{NIPS2001:2092}
A.~Y. Ng, M.~I. Jordan, and Y.~Weiss, ``On spectral clustering: Analysis and an
  algorithm,'' in \emph{Advances in Neural Information Processing Systems 14},
  T.~G. Dietterich, S.~Becker, and Z.~Ghahramani, Eds.\hskip 1em plus 0.5em
  minus 0.4em\relax MIT Press, 2002, pp. 849--856.

\bibitem{engineering_toolbox}
``Engineering toolbox. friction and friction coefficients.''

\end{thebibliography}

\end{document}